\documentclass{article}

\usepackage{arxiv}

\usepackage[utf8]{inputenc} 
\usepackage[T1]{fontenc}    
\usepackage{hyperref}       
\usepackage{url}            
\usepackage{booktabs}       
\usepackage{amsfonts}       
\usepackage{nicefrac}       
\usepackage{microtype}      
\usepackage{lipsum}		
\usepackage{graphicx}
\usepackage[square,semicolon]{natbib}
\usepackage{doi}
\usepackage{xcolor}

\definecolor{boxgrey}{HTML}{F3F3F3}

\title{Can We Trust AI Benchmarks?\\An Interdisciplinary Review of\\Current Issues in AI Evaluation
\thanks{\textbf{Disclaimer:} The views expressed in this paper are purely those of the authors and may not, under any circumstances, be regarded as an official position of the European Commission.}
}

\author{
    \href{https://orcid.org/0000-0002-7534-4268}
    {\includegraphics[scale=0.06]{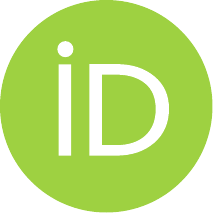}\hspace{1mm}Maria Eriksson} \\
	European Commission, Joint Research Centre (JRC) \\
	Seville, Spain \\
    \texttt{maria.eriksson@ec.europa.eu} \\
\And
    \href{https://orcid.org/0000-0002-5506-3020}{\includegraphics[scale=0.06]{orcid.pdf}\hspace{1mm}Erasmo Purificato} \\
	European Commission, Joint Research Centre (JRC) \\
	Ispra, Italy \\
    \texttt{erasmo.purificato@ec.europa.eu} \\
\And
    Arman Noroozian \\
	European Commission, Joint Research Centre (JRC) \\
	Brussels, Belgium \\
    \texttt{arman.noroozian@ec.europa.eu} \\
\And
    \href{https://orcid.org/0000-0001-6219-3977}{\includegraphics[scale=0.06]{orcid.pdf}\hspace{1mm}Jo\~{a}o Vinagre} \\
	European Commission, Joint Research Centre (JRC) \\
	Seville, Spain \\
    \texttt{joao.vinagre@ec.europa.eu} \\
\And
    Guillaume Chaslot \\
	European Commission, Joint Research Centre (JRC) \\
	Brussels, Belgium \\
    \texttt{guillaume.chaslot@ec.europa.eu} \\
\And
    \href{https://orcid.org/0000-0003-4983-3989}{\includegraphics[scale=0.06]{orcid.pdf}\hspace{1mm}Emilia Gómez} \\
	European Commission, Joint Research Centre (JRC) \\
	Seville, Spain \\
    \texttt{emilia.gomez-gutierrez@ec.europa.eu} \\
\And
    \href{https://orcid.org/0000-0003-2433-7110}{\includegraphics[scale=0.06]{orcid.pdf}\hspace{1mm}David Fernandez-Llorca} \\
	European Commission, Joint Research Centre (JRC) \\
	Seville, Spain \\
    \texttt{david.fernandez-llorca@ec.europa.eu} \\
}

\renewcommand{\shorttitle}{Can We Trust AI Benchmarks?}

\hypersetup{
pdftitle={Can We Trust AI Benchmarks? An Interdisciplinary Review of Current Issues in AI Evaluation},
pdfsubject={q-bio.NC, q-bio.QM},
pdfauthor={Maria Eriksson, Erasmo Purificato, Arman Noroozian, Joao Vinagre, Guillaume Chaslot, Emilia Gomez, David Fernandez-Llorca},
pdfkeywords={AI Benchmarks, Benchmark Critique, AI Evaluation, Safety Evaluation, AI Regulation},
}

\begin{document}
\maketitle

\begin{abstract}
    Quantitative Artificial Intelligence (AI) Benchmarks have emerged as fundamental tools for evaluating the performance, capability, and safety of AI models and systems. Currently, they shape the direction of AI development and are playing an increasingly prominent role in regulatory frameworks. As their influence grows, however, so too does concerns about \textit{how} and with what effects they evaluate highly sensitive topics such as capabilities, including high-impact capabilities, safety and systemic risks. This paper presents an interdisciplinary meta-review of about 100 studies that discuss shortcomings in quantitative benchmarking practices, published in the last 10 years. It brings together many fine-grained issues in the design and application of benchmarks (such as biases in dataset creation, inadequate documentation, data contamination, and failures to distinguish signal from noise) with broader sociotechnical issues (such as an over-focus on evaluating text-based AI models according to one-time testing logic that fails to account for how AI models are increasingly multimodal and interact with humans and other technical systems). Our review also highlights a series of systemic flaws in current benchmarking practices, such as misaligned incentives, construct validity issues, unknown unknowns, and problems with the gaming of benchmark results. Furthermore, it underscores how benchmark practices are fundamentally shaped by cultural, commercial and competitive dynamics that often prioritise state-of-the-art performance at the expense of broader societal concerns.
    
    By providing an overview of risks associated with existing benchmarking procedures, we problematise disproportionate trust placed in benchmarks and contribute to ongoing efforts to improve the accountability and relevance of quantitative AI benchmarks within the complexities of real-world scenarios.
\end{abstract}


\section{Introduction}\label{sec:intro}
Quantitative artificial intelligence (AI) benchmarks (i.e., combinations of test datasets and performance metrics that are taken to represent general or specific tasks and used to compare AI model capabilities and/or risks~\citep{raji2021}) play a central role in the release and marketing of newly developed AI models. Together with qualitative evaluation methods (such as red teaming and peer confrontations~\citep{chollet2019}), quantitative benchmarks - hereafter referred to as \textit{AI benchmarks}, or simply \textit{benchmarks} - are generally seen as providing crucial feedback signals on the performance and capabilities of AI. Indeed, they have become so critical to AI development that businesses go to great lengths to achieve good benchmarking scores, with market players like OpenAI being estimated to have spent hundreds of thousands of dollars on compute to obtain a high score at the ARC-AGI benchmark~\citep{pfister2025}.

Increasingly, AI benchmarks are also used in regulatory contexts, where the goal is to assess potential societal harms posed by AI models and systems. The most notable case is the EU AI Act \citep{AIA24}, which incorporates benchmarks in several key provisions such as high-risk assessments of AI systems, where they are expected to inform requirements on accuracy, robustness, and cybersecurity (Art. 15(2)), facilitated through AI regulatory sandboxes (Art. 58 (2)). They will also be of fundamental importance for classifying general-purpose AI (GPAI) models with systemic risks and assessing high-impact capabilities (Art. 51(1) and Annex XIII). Additionally, the first and second drafts of the Code of Practice for the EU AI Act mentions benchmarks as a risk assessment method for providers of GPAI models with systemic risks \citep{1stDraftCoP2024, 2ndDraftCoP2024}. In the US, benchmarks are also relevant in the recently revoked AI Executive Order~\citep{USAIEO2023} and the AI Diffusion Framework~\citep{USAIDiff2025}. AI benchmarks can further be expected to play a central role in the implementation of legislations such as the EU Digital Services Act (DSA)~\citep{DSA2022} and the UK's Online Safety Act~\citep{UKSafetyAct2023}, which require the largest online platforms and search engines - entities that increasingly rely on AI to curate, filter, and/or rank content to millions of users - to perform regular algorithmic audits. In short, AI benchmarks - which constitute a highly heterogenous and far from standardised set of techniques - are increasingly at the heart of policy efforts to make AI more transparent and secure. 

At the same time as benchmarks are increasingly relied upon to provide AI safety assurances, however, journalists~\citep{keegan2024, roose2024, heaven2023} and researchers in a broad range of academic fields have raised serious concerns regarding their current use. This includes critical voices being raised in fields ranging from cybersecurity~\citep{mcintosh2024}, linguistics~\citep{bowman2021}, and computer science~\citep{gema2024}, to sociology~\citep{engdahl2024}, economics~\citep{ethayarajh2021}, philosophy~\citep{lacroix2022}, ethnography~\citep{keyes2022}, and science and technology studies~\citep{keyes2022}. Such scholars have for example described current AI evaluation practices as a "minefield"~\citep{narayanan2023a}, that raise serious ethical concerns regarding what should be measured, according to what standards, and with what downstream effects~\citep{blili-hamelin2023, mitchell_how_2023, burden_evaluating_2024}. They have also emphasised that benchmarks are deeply political, performative, and generative in the sense that they do not passively describe and measure how things are in the world, but actively take part in shaping it~\citep{grill2024}. This happens as benchmarks continuously influence how AI models are trained, fine-tuned, and applied - practices with wide-ranging political, economic, and cultural effects.

With this backdrop, an interdisciplinary and up-to-date survey of quantitative AI benchmarking critique is momentously missing. The aim of this article is to address this gap by mapping known limitations in quantitative AI evaluation practices, drawing together insights from multiple academic fields. Our goal is to provide an overview of current risks associated with quantitative AI tests, targeting policy makers, stakeholders performing algorithmic audits, and developers of AI models and benchmarks. To achieve this, we gather and analyse around 110 publications published between 1\textsuperscript{st} January 2014 and 31\textsuperscript{st} December 2024 that explicitly and primarily address limitations in current benchmarking practices. Notably, more than half of these publications were published in 2023 or later, highlighting the urgency of providing an updated survey of discussions in the field.

Our findings show that a rapidly growing number of researchers are voicing concerns regarding how benchmarks are used to define and measure what is safe or unsafe, moral or immoral, true or false, toxic or healthy. They further support the notion that no benchmark is neutral and that AI tests and benchmarking practices always rest "on interwoven technical and normative decisions"~\cite[p.~1201]{rauh2024} in ways that urge policy makers and AI developers to apply them with caution. In particular, previous research highlights a need to question the relevance and trustworthiness of well-cited benchmarks, since existing studies have repeatedly shown weaknesses in benchmarks perceived as state-of-the-art (SOTA). Our findings also reveal a substantial need to scrutinise capability and safety-oriented AI benchmarks to the same extent as the AI models they are meant to evaluate. In short, AI benchmarks need to be subjected to the same demands concerning transparency, fairness, and explainability, as algorithmic systems and AI models writ large.

In what follows, we first provide a background discussion on AI benchmarking practices, including clarifications on the terminology used in this paper. Next, we situate our work within a series of previous research that summarise benchmark critique, before providing a summary of nine issues with quantitative benchmarks, identified during the course of our research. While not exhaustive, we present these issues as a taxonomy of benchmark critique that highlights crucial and often interrelated points of concern regarding benchmarks, voiced in the past decade. Finally, we provide some concluding remarks on the implications these issues have for policy makers. On the whole, our work constitutes the first step in a broader research and policy-oriented project aimed at developing a framework for trustworthy AI benchmarks, and will proceed by considering mitigation strategies.

\section{Background}\label{sec:background}
Etymologically, the term ``benchmark'' has its roots in land surveying, where a physical mark (known as a ‘bench mark’) was used as a reference point for measuring elevations. This mark typically consisted of a horizontal groove in a surface, which supported a level surface or 'bench' for a levelling rod. Over time, the term evolved to encompass a broader meaning, referring to any standard or reference point used for comparison or evaluation~\citep{oed}. 
Benchmarking is currently applied across many different domains such as bioinformatics~\citep{Aniba2010}, environmental quality~\citep{Henning2000}, security~\citep{noroozian_evaluating_2020}, information retrieval~\citep{thakur2021beir}, transistor hardware~\citep{Cheng2022}, industry and business~\citep{Camp1989}, and within the public sector~\citep{Bruno2014}, where it often refers to procedures for comparing the performance or best practices of different companies or processes.
Here, we focus on the use of benchmark tests within computing, where they are used to evaluate the performance of hardware or software systems by comparing them to a standard or reference point \citep{Henning2000}. More specifically, we zoom in on AI development, where benchmarks often facilitate cross-model comparisons, track model progress, and identify weaknesses \citep{reuel_betterbench_2024}.  

Benchmarks are often perceived as comparably cost- and time-effective tools for AI providers who may run them regularly throughout the model development to obtain signals of model capabilities \citep{weidinger_sociotechnical_2023}. They can be applied to both software and hardware solutions, where the latter, for instance, evaluates the performance of CPUs, GPUs, or TPUs \citep{Mattson2020}, and/or hardware accelerators \citep{Mattson2020b}. We address software-oriented benchmarks. Based on the definition proposed by \citet{raji2021}, we define such benchmarks as “a particular combination of a set of test datasets, including human-in-the-loop interactions, and associated performance metrics, conceptualised as representing one or more specific tasks or related capabilities, typically chosen by a community of researchers as a shared framework for comparing AI models”. A \textit{testing dataset} would refer to “a separate dataset, distinct from the training data, used to objectively evaluate the performance and generalisability of an AI model”. These datasets are typically composed of \textit{samples} or \textit{instances} that include an input paired with the desired output (also known as reference, gold, or ground truth). A \textit{task} would be “a particular specification of a problem, typically represented as a mapping between an input space and an output or action space, extensionally defined through one or more datasets, and usually associated with a specific performance evaluation metric” (based on \citep{raji2021} and \citep{schlangen_targeting_2020}). And finally, a \textit{metric} would refer to "a specification of the mechanism to determine the degree of success or failure of the model's outputs". The metric represents a way to summarise model performance over a given task and dataset, usually defined as a single number or score \citep{raji2021}. AI models that achieve the best scores on the metrics of a benchmark are generally considered SOTA in terms of performance. 

While humans are always involved in the design and creation of ground truth in benchmarks, they can play a more or less direct and active role in applying benchmark tests. In \textit{automated} or \textit{quantitative benchmarks}, a set of tasks, datasets, and metrics are first defined through human decision making. The execution of a benchmark test is then carried out without direct human intervention. In \textit{qualitative} benchmarks, humans intervene and partake in evaluations, for example, as evaluators, judges, or real-time interrogators (e.g., adversarial testing or red teaming). We primarily consider quantitative benchmarks. This is not to say that there is no relevant critique concerning qualitative AI evaluations, but such methods introduce a different set of problems that is out of scope for our discussions here.

\section{Related Work}\label{sec:relatedwork}
Over the years, several surveys and meta-reviews have set out to summarize discussions on limitations in AI benchmarking and this paper represents a continuation of such efforts. In 2021, \citet[p.~1]{liao2021} identified a wide range of "surprisingly consistent critique" directed at benchmarking practices across fields such as computer vision, natural language processing, recommender systems, reinforcement learning, graph processing, metric learning, and more. The authors present a taxonomy of observed benchmark failure modes, including implementation variations, errors in test set construction, overfitting from test set reuse, and comparisons to inadequate baselines. Along similar lines, \citet{hutchinson_evaluation_2022} survey discussions on evaluation practices in machine learning, particularly in the fields of computer vision and natural language processing. The authors identify eight key evaluation gaps, including topics such as "neglect of model interpretability" and "oversimplification of knowledge" and argue for a shift toward application-centric evaluations that account for safety, fairness, and ecological validity. Similar points are also raised by \citet{gehrmann2023}'s most recent meta-review, which categorizes issues with text-oriented benchmarks over the past two decades, and propose a long-term vision for improving evaluation practices, emphasizing the need for comprehensive evaluations that include multiple datasets, metrics, and human assessments. Their key recommendations include focusing on model limitations, enhancing dataset documentation, and adopting a more nuanced approach to evaluation to better characterize model capabilities.

Our review revisits these areas of critique and traces how debates concerning benchmarks have evolved in the last few years due to rapid developments in the field. Recent research has identified an unprecedented growth in the release of AI benchmarks, especially in the area of safety, starting from 2023 and onwards~\citep{rottger2024}. Our research indicates a similar increase in publications expressing benchmark critique. For instance,~\citet{rottger2024} found that roughly 46\% of the AI safety benchmarks they identified as relevant in their survey of the field had been produced in 2023, with as many as 15 new datasets being released during the first two months of 2024. Likewise, almost 55\% of the articles we identified as relevant during the course of our research had been released in 2023 or later.
Given this rapid increase in attention given to the topic, our primary goal is to extend existing meta-reviews into the present. Whereas previous meta-reviews have tended to focus on critique raised in a limited number of domains (primarily natural language processing and machine learning) we also include interdisciplinary accounts from the humanities and social sciences in our review. Furthermore, we provide insights concerning more recently identified areas of benchmark concern, such as works revealing that AI models can be programmed to underperform on benchmarking tests (a problem known as ``sandbagging'')~\citep{weij2024}, and studies presenting evidence that AI safety benchmarks strongly correlate with upstream AI capabilities (causing reasons to worry about AI ``safetywashing'') \citep{ren2024}, insights that both raise additional questions about the validity and trustworthiness of current benchmarking practices.

\section{Methodology}\label{sec:methodology}
When gathering source materials for this review (and instead of conducting systematic keyword searches in research databases) we opted for a snowball sample~\citep{jalali2012systematic,badampudi2015experiences}, which helped us identify papers that \textit{primarily} address benchmark \textit{critique} (our main inclusion criteria) while avoiding noise in the form of articles that present new benchmarks or simply apply them (our main exclusion criteria). A snowball method also helped us cut across interdisciplinary inconsistencies in terminology and allowed us to discover pre-prints - a publication form where much debate concerning benchmarks has taken place, but that would have been missed in a traditional review based on scientific databases. We initially started from the article "AI and the Everything in the Whole Wide World Benchmark" \citep{raji2021} and expanded by reviewing articles included in its reference list. We also studied how the text has been cited since its publication using Google Scholar - a process that was repeated for each article considered relevant. When important concepts (such as sandbagging) appeared in the literature, targeted searches for research in these fields were made. We also made special efforts to include discussions concerning benchmarks for images, sound, and audiovisual content which are under-represented compared to text-focused benchmarks.

In the process, we surveyed a wide range of papers that, for instance, highlight general problems with the production of AI datasets \citep{mitchell2019,orr2024b}, propose alternative ways of designing benchmark leaderboards \citep{rodriguez2021,liu2021}, or discuss the wider use of proxies in tests and evaluations \citep{mulvin2021,pinch1993,marres2020}.
Such papers provide important contextual insights to discussions concerning benchmarks. However, they were omitted from what we considered our core collection of previous research. This collection ended up consisting of about 110 papers and articles that \textit{explicitly} and \textit{primarily} highlight issues with benchmarks. Notably, articles that propose new benchmarks were \textit{not} added to this collection by default, even though such articles naturally contain some level of benchmark critique. The reasoning behind this is twofold. On the one hand, it was necessary to limit the size and scope of our review. On the other hand, many articles proposing new benchmarks (if not most of them) do not question or discuss many (if any) of the underlying assumptions that benchmark assessments rely on.
In that sense, they reproduce the general notion that quantitative benchmarks provide a reasonable technical "fix" to issues with AI safety and capability assessments, rather than raising more fundamental and critical discussions on limitations in benchmark use and design. For these reasons, they were also excluded from our core collection of papers. Our collection was further limited by only containing articles published between 1\textsuperscript{st} January 2014 and 31\textsuperscript{st} December 2024. While it is true that AI models have been tested and evaluated since their origins in the mid 1950's (with the Turing test serving as an iconic example), we limit our review to this decade long period, since 2014 marks the starting point for recent intensifications in AI research and development. 

Our resulting meta-review is not exhaustive but it covers a broad range of critique that has been aimed at benchmarking practices. The nine issue categories presented were identified after close-reading and relevant articles, classifying and grouping papers voicing similar critique, and discussing these classifications within the wider author group. The resulting issue areas represent the result of these discussions, although many issues were difficult to categorize. E.g., problems with Anglo-centric benchmarking datasets could simultaneously be framed as a cultural problem (Anglo-centrism), a construct validity problem (as English is taken to represent culture as a whole), and a problem with narrow benchmark diversity and scope. As a result of such overlaps, identified issues are not meant to be read as an absolute and all-encompassing definition of benchmark vulnerabilities, but as a narrative tool to present our findings. We especially focus on works that voice critique with relevance for policy makers and policy implementation, highlight areas of concern that cut across different modalities (text, images, sound, moving images), and point toward fundamental weaknesses in the design and application of benchmarks, as opposed to research that critiques individual benchmarks. Our goal has further been to provide a \textit{diverse} account of concerns regarding benchmarks.

\section{Nine Reasons to Be Cautious with Benchmarks}\label{sec:issues}
In the sections below, we summarise the main issues that were identified during the course of our research. Importantly, these issues - presented here as a taxonomy - are not arranged according to their importance or urgency. They are also not meant to be understood as isolated issues, but rather deeply interlinked problems. Indeed, this complexity and interdependence is precisely what makes AI evaluations challenging.

\subsection{Problems with Data Collection, Annotation, and Documentation}

An initial set of issues with AI benchmarks found during our research is limitations in the collection, annotation, and documentation of benchmark datasets. This ties into a broader critique regarding insufficient documentation in AI research which is central to calls for more transparent and trustworthy algorithmic systems~\citep{gebru2021, mitchell2019, orr2024b, simson2024, scheuerman2021}. Research has found that it is often difficult to trace precisely \textit{how}, \textit{when}, and by \textit{whom} benchmark datasets have been made \citep{reuel_betterbench_2024, denton2020} which compromises the ability for benchmarks to be used in robust and generalisable ways \citep{arzt2024}. The issue has partly been linked to the low status of dataset-related work within the machine learning community (which instead privileges model development) \citep{orr2024, sambasivan2021}, and the fact that AI datasets are often “reduced, reused, and recycled” \citep{koch2021}, which complicates documentations of their possible limitations \citep{thylstrup2022, park2022}. Notably, \citet{koch2021} have found that more than 70 percent of the benchmark datasets used in prominent computer vision papers had been reused from other domains.
\citet{park2022} also find that while benchmark datasets often contain ample information on how to use the dataset, documentation is often missing concerning their shortcomings and social impact. 
In their exploration of benchmark sharing platforms like HuggingFace\footnote{\url{https://huggingface.co/}. Last seen \today.} and PapersWithCode\footnote{\url{https://paperswithcode.com/}. Last seen \today.} the same researchers also note that confusing metadata terminology severely complicates efforts to understand benchmark dataset documentation. 

Like AI training datasets, benchmarks have been singled out for raising ethical and legal questions concerning copyrights, privacy, informed consent, and rights to opt-out \citep{paullada2021}. Many benchmarks rely on crowd-sourced or user-generated content from platforms like Wikihow\footnote{\url{https://www.wikihow.com/}. Last seen \today.}, Reddit\footnote{\url{https://www.reddit.com/}. Last seen \today.}, or trivia websites \citep{keegan2024, grill2024}, whose annotation may be biased, lack input from expertise in specialised fields, or be produced under exploitative conditions \citep{tsipras2020, aroyo2015, sen2015}. Previous research has noted that this matters when baselines for "good", "bad", and "safe enough" AI models are calibrated in sensitive contexts \citep{rauh2024, grill2024}. Moreover, the absence of human performance references and difficulty rubrics in benchmarks has been highlighted, which are increasingly considered important factors in evaluating capabilities and generality \citep{chollet2019}.

Noisy human annotations and lack of care in the making of benchmark datasets have further been found to skew benchmark scores \citep{kejriwal_noise_2024} and result in AI models exploiting unknown quirks and spurious cues in the training data, rather than solving their original intended task \citep{liao2021, paullada2021, geirhos_shortcut_2020}.
For instance, \citet{oakden-rayner2019} found that an X-ray image classification model predicted collapsed lungs with high accuracy, although it turned out the model only identified the presence of a chest drain (used to cure the condition) that was (unknowingly) present in a majority of the positive training images. In other words, the model performed well in benchmark tests, but the reason why completely sidestepped the original purpose of the task (identifying collapsed lungs). When the all images containing chest drains were removed from the training dataset, the model performance dropped by over 20\%.
Similar patterns have also been confirmed in recent research on LLM evaluation \citep{pacchiardi2024} and is sometimes discussed as failure in the "world model" of an AI system \citep{vafa2024}.
While it could be argued that picking up on spurious cues is a sign of intelligence and therefore a capability worth noting, the issues above highlight how a lack of in-depth attention to the data that benchmarks rely on - alongside a failure to acknowledge how little is actually known about \textit{how} and \textit{why} AI models perform well in benchmark tests - can produce frail and uncertain AI evaluations.

\subsection{Weak Construct Validity and Epistemological Claims}

Another genre of benchmark critique focuses on the epistemological claims that tend to surround benchmarks and examines the limits of what can be known through quantitative AI tests. A central reference point in these discussions is the observation by \citet{raji2021} that many benchmarks suffer from construct validity issues in the sense that they do not measure what they claim to measure. As the authors proclaim, this is especially troublesome when benchmarks promise to measure universal or general capabilities, since this vastly misrepresents their actual capability. As a result, the authors argue that framing an AI benchmark dataset as general purpose "is ultimately dangerous and deceptive, resulting in misguidance on task design and focus, underreporting of the many biases and subjective interpretations inherent in the data as well as enabling, through false presentations of performance, potential model misuse".~\cite[p.~5]{raji2021}. At the heart of this critique lies the realization that many benchmarks do not have a clear definition of what they claim to measure, which makes it impossible to measure if they succeed in the task or not \citep{blodgett2021, bartz-beielstein2020}. In a close analysis of four benchmarks used to evaluate fairness in natural language processing (StereoSet, CrowS-Pairs, WinoBias, and WinoGender), \citet{blodgett2021} for example found that all benchmarks revealed severe weaknesses in terms of defining what is being measured. For instance, culturally complex and highly contested concepts like "stereotypes" or "offensive language" were left unspecified, causing a series of logical failures and interpretational conflicts. Elsewhere, research has shown strong disagreements in how benchmark tasks are \textit{conceptualised} and \textit{operationalised }\citep{subramonian2023}, and found that benchmarks are applied in highly idiosyncratic ways \citep{rottger2024} that can be misleading in numerous ways~\citep{leech_questionable_2024}.
Frequently, the difficulty in defining what benchmarks evaluate persists since there is no clear, stable, and absolute ground truth for what is claimed to be measured \citep{narayanan2023}. Since concepts like "bias" and "fairness" are inherently contested, messy, and shifting, benchmarks that promise to measure such terms will inevitably suffer from an "abstraction error" that produces a false sense of certainty \cite[p.~63]{selbst2019}. 

It has also been pointed out that many benchmark datasets are inadequate and/or useless proxies for what they are meant to evaluate. For instance, researchers have identified a slippage in distinguishing between algorithmic "harms" and algorithmic "wrongs" when evaluating the capabilities of AI models - two not necessarily overlapping concepts \citep{diberardino2024}. Others have questioned whether the content of benchmark datasets is a reasonable substitute for the "real world" scenarios they are meant to reflect. For instance, the decision to use examples generated by Amazon crowdworkers and posts from the Reddit forum "\textit{Am I the asshole?}" as proxies for ethics and morals in benchmarks such as HellaSwag - a widely cited benchmark for language models has been questioned \citep{keegan2024}. In benchmarks consisting of professional exams, researchers have further argued that such tests "emphasize the wrong thing" and "overemphasize precisely the thing that language models are good at" and are thus unreliable measures of things such as medical or legal skill \cite[n.p]{narayanan2023}.
As \citet{narayanan2023} put it, "it's not like a lawyer's job is to answer bar exam questions all day". 
A recent study by \citet{ren2024} also found that many widely used safety benchmarks (including ETHICS, TruthfulQA, GPQA, QuALITY, MT-Bench, LMSYS Chatbot ARENA, ANLI, AdvGLUE, and AdvGLUE++) 
highly correlate with general and upstream model capabilities, raising concerns regarding “safetywashing” as applying them could imply that “capability improvements are misrepresented as safety advancements". While it could be argued that capability and safety are largely entwined (the more capable a model is, the higher the likelihood it could cause harm), \citet[p.~1]{ren2024} suggest that a blurred distinction between the two may hide the fact that severe biases and safety issues in AI models can persist, even as their overall capabilities improve.

\subsection{Sociocultural Context and Gap}

Another key insight from previous research concerns the importance of the social, economic and cultural contexts where AI benchmarks are created, used, and maintained. Among researchers engaging in benchmark critique, we identify a strong consensus regarding the need to recognise that benchmarks are ultimately "normative instruments that perpetuate particular epistemological perspectives about how the world is ordered" \cite[p.~1877]{orr2024a}. Qualitative research has also examined the cultural and social environments where benchmarks are made, finding that they are deeply shaped by shared and arbitrary assumptions, commitments, and dependencies \citep{engdahl2024,michael2022, scheuerman2021,orr2024,sambasivan2021,paullada2021}.
Such assumptions for example include valuing "efficiency at the expense of care; universality at the expense of contextuality; impartiality at the expense of positionality;
and model work at the expense of data work" \citep{scheuerman2021}, or reproduce contested ideas such as the notion that low-quality data can be drowned out by scale \citep{orr2024}.
Recent research also illustrated that AI safety research and benchmark competitions are increasingly informed by political movements and ideologies such as longtermism and effective altruism~\citep{ahmed2024}.

Scholars have further identified sociotechnical gaps and a lack of consideration for downstream utility as a concern in AI benchmarking \citep{hutchinson_evaluation_2022,frieder_data_2024}.
\citet{liao2023} highlight that this means it is often unclear who is meant to care about benchmark evaluation results and how they should be used in practice.
For instance, a recent study by \citet{blagec2023}, which compared the explicitly stated needs for AI technologies among clinical medical practitioners with existing clinical benchmark datasets, found that most benchmarks failed to answer to the needs of medical experts.
They also found that benchmarks for the most urgently requested medical or clinical tasks were completely missing and noted that similar misalignments likely exist in many other fields. For example, \citet[p.~79]{jannach2020} highlight a lack of attention to how recommender systems "may create value; how they, positively or negatively, impact consumers, businesses, and the society; and how we can measure the resulting effects". \citet{ethayarajh2021} further argue that failures to consider the practical utility of benchmarks has for example made it possible to ignore the discriminatory and environmental damages of AI technologies and allowed for highly energy-inefficient and deeply biased AI models to reach the top of most benchmark leaderboards.

\subsection{Narrow Benchmark Diversity and Scope}

Previous research has further found that current benchmarking practices suffer from diversity issues, a problem that is also found within the broader AI ecosystem  \citep{gomez2024}. On the one hand, a vast majority of benchmarks focus on text, while other modalities (audio, images, video, and multimodal systems) remain largely unexamined \citep{rauh2024, weidinger_sociotechnical_2023, rottger2024}. This concentration is problematic since AI models are increasingly multimodal in scope. \citet{guldimann2024} also find that benchmarks addressing user privacy, copyright infringement, and interpretability are currently lacking and incomprehensive, while benchmarks in safety areas dealing with corrigibility and explainability are practically missing all together. This leads the authors to conclude that current practices for evaluating topics like safety and ethics “are often simplistic and brittle, leading to inconclusive results” \cite[p.~3]{guldimann2024}.
\citet{koch2021} further identify a concentration on fewer and fewer benchmark datasets within most task communities and note that dominant benchmarks have been introduced by researchers at just a handful of elite institutions, raising questions about representation diversity in the design of benchmarks. Scholars have also found that current AI safety evaluation practices almost exclusively deal with English content \citep{mcintosh2024, rottger2024, poelman_roles_2024}, and are frequently based on datasets where minorities are under-represented, despite efforts to diversify them \citep{simson2024}. This raises concerns regarding the inclusion of multiple perspectives on complex topics like ethics and harm. 

Aside from mainly focusing on a small range of tasks, previous research has identified that most benchmarks tend to be abstracted out of their social and cultural context \citep{selbst2019, lum_bias_2024}, are aggregated in problematic ways \citep{burnell_rethink_2023}, and rely on a static, one-time testing logic, where results from single evaluations are taken to represent model capabilities writ large. This has given rise to calls for more multi-layered~\citep{weidinger_sociotechnical_2023}, longitudinal~\citep{mizrahi2024}, and holistic evaluation methods that can be reproduced~\citep{kapoor_ai_2024}, and prove that AI models do not just perform well in controlled environments, but also in critical/real-world circumstances over time \citep{ojewale2024, mcintosh2024, chang2023}. In their survey of LLM benchmark inadequacies, McIntosh et al. note that "current benchmarks generally adopt task-based formats, such as Multiple Choice Questions (MCQs) and dialogue-based evaluations, which tend to be static and do not capture the evolving nature of human-AI interactions" \cite[p.~6] {mcintosh2024}.
\citet{reuel_betterbench_2024} further note that failures to re-run evaluations multiple times (using different random seeds and sampling temperatures, for example) imply that very little is known about the intra-model variance of benchmarks.
As a result, they conclude that it is possible that "most benchmarks fail to distinguish signal and noise" \cite[p.~9]{reuel_betterbench_2024}. 
Others have emphasised that AI audits often fail to consider risks associated with multiple (inter)acting AI systems \citep{birhane2024}, and rarely take human actions and motivations into consideration \citep{weidinger_sociotechnical_2023, chang2023, rauh2024}.

While most benchmarks are designed to tell us something about a model's success, research has further pointed out that they often reveal little (or nothing) about their particular ways of \textit{making mistakes}, which is crucial from an AI safety and policy enforcement perspective. As \citet{gehrmann2023} put it, "ranking models according to a single quality number is easy and actionable - we simply pick the model at the top of the list - [yet] it is much more important to understand when and why models fail" \cite[p.~130]{gehrmann2023}.
For instance, they suggest that a focus on errors and fragilities (as opposed to instances of success) can be useful for developers of smaller models, since "work on quantifying shortcomings is equally applicable to smaller models and methods that improve model robustness often work on all model sizes" \cite[p.~131]{gehrmann2023}.
In this sense, failure-focused benchmarks could play an important role in equalling out the playing field in AI development.

\subsection{Economic, Competitive, and Commercial Roots}

Another contextual element that has been singled out as important is the competitive and commercial roots of benchmark tests.
Previous research has emphasised that capability-oriented benchmarks are deeply embedded in corporate marketing strategies and play an important role in increasing the AI hype, attracting customers and investors, and showcasing how models outperform competitors \citep{orr2024a, grill2024, Zhijia2024}. As \citet[p.~1881]{orr2024a} put it, benchmarks "serve as the technological spectacle through which companies such as OpenAI and Google can market their technologies". Many benchmarks also have origins from within the industry and are capability-oriented and centred around tasks with a high potential economic reward, as opposed to focusing on other goals such as ethics and safety \citep{ren2024, ethayarajh2021}.

Previous research has noted that this competitive and corporate embedding discourages thorough self-critique since there is a direct "incentive mismatch between conducting high-quality evaluations and publishing new models or modelling techniques" \cite[p.~103]{gehrmann2023} and that the field of AI development is "turning into a giant leaderboard, where publication depends on numbers and little else (such as insight and explanation)" \citep{church_survey_2019}. 
While a lack of incentives to disclose weaknesses and limitations is a general problem in science \citep{smith2022}, it has been pointed out that benchmarks have played an especially central role in naturalizing and solidifying a competitive culture in AI research, which is increasingly approached as a "sport" \citep{orr2024a}. As of late, benchmark evaluations have also become increasingly professionalized and transformed into an industry in itself with the rise of platforms like Kaggle and Grand Challenge, who provide organisational and infrastructural support to AI competitions and increasingly function as infrastructures of power in fields like medical imaging \citep{luitse2024}. The issue of optimising for high benchmark scores at the expense of insight and explanation is known as a form of SOTA-chasing \citep{koch2021} and sometimes described as the "benchmark effect" \citep{Stewart2023}. Previous research has also described benchmarking as an example of what \citet{stengers2018} calls "fast track research" which idolises rapid, cumulative publication \citep{maleve2023}, thus producing a "winners curse" in AI development \citep{sculley2018}. 

Risks associated with the competitive and commercial roots of benchmarks have further been linked to the growing influence of industry in AI research, where private businesses share of the biggest AI models has increased from 11\% in 2010 to 96\% in 2021 \citep{ahmed2023}. In such a context, researchers have noted that current benchmarking tasks - which are generally highly data-intensive - are especially well suited to fit AI models that have been developed within the industry, whose access to advanced data infrastructures, computing power, valuable datasets, and skilled researchers now vastly exceed those of academic researchers \citep{ahmed2023}.
This concentration of power could potentially stifle robust AI evaluations and hinder the development of AI models that adhere to other aims and goals than commercial ones. Scholars have also warned that if academic researchers continue to uphold data-intensive benchmark tests as the SOTA, there is a risk that their research will become increasingly dependent on technological infrastructures provided by the industry \citep{koch2024}.

\subsection{Rigging, Gaming, and Measure Becoming Target}

A closely related issue concerns how benchmark tests can be tricked and gamed. In areas and modalities where best-practice benchmarks are missing (i.e., practically all modalities, except for text-based benchmark evaluations), researchers have noted that there are strong incentives to "rig" benchmark tests. For instance, \citet{dehghani2021} survey how know-how and recipes for how to score high on benchmark setups are often widely circulated online.
Recently, language models have also been found to be optimised for answering the multiple choice questions that are often part of benchmarks \citep{alzahrani2024}, and to (either intentionally or unintentionally) "fake" alignment with ethics or safety goals \citep{greenblatt2024} and hide their true capabilities and objectives - also known as \textit{scheming} \citep{meinke_frontier_2024}.
The issue points towards what is known as Goodhart's law: ``when a measure becomes a target, it ceases to be a good measure''~\citep{strathern1997}.

One reason why gaming can proceed is that users of benchmarks rarely provide the resources needed to validate and replicate their test results \citep{bartz-beielstein2020, dehghani2021, biderman2024, reuel_betterbench_2024} and that they rely on black-box (as opposed to white-box) access to AI models, which undercuts rigorous audits \citep{casper_black-box_2024}.
To examine the relevance and validity of benchmark scores, researchers have further emphasised that it is necessary to access information concerning all aspects of the evaluation procedure, such as the original evaluation code and all details concerning the experimental setup) \citep{biderman2024}. Providing such documentation is far from standard, however, especially when proprietary AI models are concerned. This makes it possible to tweak and cherry-pick benchmark results - a problem that is especially pressing given that subtle "variations in prompts, formatting or other implementation details can significantly impact the performance and validity of evaluations" \cite[p.~3]{biderman2024}. In their in-depth analysis of 24 SOTA language model benchmarks, \citet{reuel_betterbench_2024} found that only four provided scripts to replicate the results and that no more than ten performed multiple evaluations or reported the statistical significance of their results. 

An increasingly well discussed issue also concerns the problem of "data contamination" i.e., the risk that the models have either intentionally or unintentionally ingested benchmark datasets during training, which severely questions the integrity of AI tests \citep{xu2024, zhang2024, besen2024, magar2022, roberts2023, yang_rethinking_2023}. The problem - which can be an instance of data leakage \citep{kaufman_leakage_2012, xu_benchmarking_2024} or train-test-overlap \citep{lewis2021} - has been known for long, and produces similar effects to those of overfitting and memorization \citep{tirumala2022,magar2022}, leading to models with low generalization power that perform well on familiar tasks (in-distribution) but fail other tasks with a similar difficulty and distribution shift (out-of-distribution) \citep{yuan2023, xu2024, zhang2024, besen2024, magar2022, roberts2023, narayanan2023}. When testing GPT4 on benchmark problems from Codeforces (a website hosting coding competitions) in 2023, for instance, \citet{narayanan2023} found that the AI model could regularly solve benchmark problems classified as easy - as long as the problems had been added before 5\textsuperscript{th} September 2021. For problems added later, GPT4 could not get a single question right, suggesting that the model had memorised questions and answers. Similar results have also been identified in multiple other models and benchmarks \citep{xu2024, zhang2024, besen2024, magar2022, roberts2023}. 
Despite the fact that issues with data leaks are so well known that strategies have been developed to avoid it\footnote{\url{https://livebench.ai}. Last seen \today.}, there is still a widespread lack of reporting of data contamination tendencies during benchmark tests. In a study from October 2024, for example, \citet{zhang2024} found that out of 30 analysed models, only 9 reported train test overlap. 

Another recently identified issue is called "sandbagging" and involves a "strategic underperformance on an evaluation" which occurs when an AI developer intentionally \textit{understates} a models capability, for instance to avoid becoming a target for AI safety regulation. In a 2024 study, for example, \citet[p.~1]{weij2024} prompted frontier models like GPT-4 and Claude 3 Opus and found that they "selectively underperform on dangerous capability evaluations, while maintaining performance on general (harmless) capability evaluations". The researchers also found that that it was possible to fine-tune and adjust both frontier and smaller models to hide specific capabilities or target specific capability scores. The issue could be generalised to popular benchmarks like the Weapons of Mass Destruction Proxy Benchmark (WMDP), and puts the trustworthiness of benchmark evaluations into question, especially in a regulatory context.

\subsection{Dubious Community Vetting and Path Dependencies}

A related area of research emphasises how benchmarks become naturalised and reach standard status because of the culture and logic of academic citations \citep{orr2024a}. For instance, new benchmarks are commonly introduced together with new or updated AI models. If the AI model becomes popular, the benchmark may become widely cited and circulated as a secondary effect, even though the developers of the benchmark did not intended or expect it to become standard. In this way, a benchmark that might have been selected for practical purpose or due to subjective preferences may come to have substantial technical and cultural influence, even though its suitability as a yardstick for ethics, safety, or performance may be questioned \citep{orr2024a, orr2024}. For instance, \citet{denton2021} show how this was partly the case when the ImageNet dataset - a key reference point in the performance testing of computer vision models - became standard, following  the unforeseen success of the ImageNet Large Scale Visual Recognition Challenge hosted by Toronto University in 2010. Likewise, the so-called Lena test image - also central to computer vision benchmark tests - was taken from the centerfold of a November 1972 Playboy magazine and catapulted into computer history since "someone happened to walk in with a recent issue of \textit{Playboy}" at a time of need, according to what has been described as the most credible origin story \cite[p.~80]{mulvin2021}.
Since then, the ImageNet dataset and its associated benchmark challenge has been become a symbol for dataset bias \citep{denton2021}, while the Lena test image has come to serve as a prime example of the role of whiteness and women's sexualised bodies in the standardisation of digital visual culture \citep{mulvin2021}. These examples highlight how benchmarks are fundamentally cultural and political products whose power and influence may and be (uncritically and problematically) reinforced through community vetting. 

Aside from bias and representational issues, \citet[p.~1]{schlangen_targeting_2020} argues that what is typically missing and left implicit in the peer-review-fuelled process of benchmark use is the argumentation for why scoring well on a particular benchmark "constitutes progress, and progress towards what". Instead, researchers are expected to routinely demonstrate performance on dominant benchmarks, despite the fact that more task-specific benchmarks may be more technically appropriate \citep{koch2021}. In this way, peer-washing serves to "maintain datasets as authoritative proxies even when they are shown to be harmful or problematic" \cite[p.~4966]{orr2024}. New benchmarks often have a difficult time to gain traction because of the dominance and authority of well-cited benchmarks \citep{jaton2021}, and in a study of 3765 benchmarks, \citet{ott2022} found that only a small minority of the proposed benchmark solutions reached widespread adoption. This problem is also recognised within the industry. For instance, researchers at Google's Brain Team describe what they call a "benchmark lottery" which "postulates that many factors, other than fundamental algorithmic superiority, may lead to a method being perceived as superior" \cite[p.~1]{dehghani2021}. Ironically, researchers have simultaneously found that a majority of influential benchmarks have been released as preprints without going through rigorous academic peer-review \cite[p.~6]{mcintosh2024}. In other words, problems with peer-washing both concern the poor quality control that many benchmarks are subjected to, and the self-fulfilling and sometimes excessive ways through which certain benchmarks are propelled into standards.

\citet{bao2022a} further note that most papers that introduce benchmarks have origins in the field of machine learning and are mainly focused on methods. This means that the content of benchmarks datasets are often considered secondary, since the datasets are merely there to provide a baseline for comparison between different evaluation methods. When benchmarks are applied to real and specific use-cases - such as evaluating the fairness of algorithmic risk assessment instruments within the criminal justice system - the downstream effects of such a lack of concern for datasets can have worrying effects.
As \citet{bao2022a} put it, a paper on benchmarks can be of high quality in a pure AI/ML methods sense, but irrelevant, dangerous, or harmful when applied in circumstances such as the criminal justice system, since it may introduce and perpetuate harms and mistranslations. The researchers also suggest that the current peer-review system implies that benchmarks that are primarily relevant from a methods/machine learning perspective will be cited far more often than benchmarks that are relevant for use in specific, real-life use-cases. Looking too closely at citation counts when determining the quality and relevance of benchmarks may thus effectively lead users astray and complicate identifying benchmarks with a high \textit{practical} utility.

Problems with dubious community vetting become especially worrying given that benchmarks create "path dependencies" in AI research, meaning they reinforce certain methodologies and research goals, while stifling those that do not align with the logic of dominant benchmark tests \citep{blili-hamelin2023}.
In particular, \citet[p.~3]{koch2024} warn against the tendency for benchmarks to favour a form of task-driven scientific monoculture that privileges immediate, explicit, formal, quantitative, and easily-interpretable evaluation mechanisms that "prioritize one or a few key epistemic values (e.g., accuracy, safety)," at the expense of a broader and more complex vision of scientific progress. The authors note how a series of methodological paradigms in AI research (such as boosting, Bayesian networks, Bayesian non-parametrics, and support vector machines) have rapidly faded away following the boom in deep learning research around the year of 2014, and partly attribute this to the "epistemic narrowness" of current benchmark tests, which are well-suited for deep learning models (whose performance can be reliably increased through scaling), but less favourable for other methodological paradigms in AI. "When scientists can only gain high-status publications by demonstrating SOTA accuracy," writes \citet[p.~30]{koch2024}, "the safest research choice becomes incrementally advancing proven methods, not innovating new ones". According to Koch, the cost of current benchmark practices - which privileges scale, compute, and the use of larger and larger training datasets - for instance includes repeated privacy and copyright violations, emotional harms caused to under-paid data workers, and increasing ecological and environmental pressures due to high energy consumption. 

\subsection{Rapid AI Development and Benchmark Saturation}

Another social, economic, and cultural issue is the speed of AI developments. As the capabilities of AI models have increased manifold in the past decade, researchers have emphasised that many benchmarks are old and designed to test models far simpler than those in use today \citep{biderman2024}. For instance, \citet[p.~5]{biderman2024} find that many prominent LLM benchmarks (including Lambada, AI2 ARC, OBQA, Hella Swag, and WinoGrande) were "designed prior to shifts such as in-context learning and chat interaction, and therefore were not designed to take these formats and approaches into account", noting that this may affect their validity in unforeseen ways. 
Many benchmarks also struggle with the challenge of quickly being outperformed as AI models achieve 100 percent accuracy scores in tests \citep{hendrycks2021, bowman2021}. A look at current AI safety leaderboards such as HELM also reveal that many AI models score notably high on benchmarks that are widely applied today. The tendency for AI models to outperform benchmarks has been described as an issue of \textit{saturation} and implies that a benchmark no longer reflects model performance \citep{ott2022}. 

Relatedly, \citet{mcintosh2024} emphasise that many benchmark frameworks are slow and complicated to implement, meaning evaluation processes can span weeks or months, which hinders timely feedback on safety risks. This becomes an issue since new model releases often enter markets continuously, making it difficult to relocate evaluation resources in quick and adequate ways. It also undermines a "benchmark's ability to consistently evaluate reasoning, comprehension, or multimodal integration, as the results may vary with each model iteration" \cite[p.~13]{mcintosh2024}, which is especially concerning in a regulatory setting, where quick, fair, and accurate AI assessments are key.
The use of thresholds, either on benchmark performance or training compute (i.e., the amount of computational resources used to train an AI model), is often seen as central to determine which AI models should warrant further regulatory scrutiny \citep{AIA24, USAIEO2023, USAIDiff2025}. However, several limitations have been identified in relation to such attempts. Creating benchmarks that can keep pace with the rapid development of AI is increasingly challenging, at the same time as recent approaches (such as auxiliary or bootstrapped models or post-training interventions) enable enhanced capabilities with reduced training compute \citep{hooker2024}.

\subsection{AI Complexity and Unknown Unknowns}

A final issue that has been discussed in relation to benchmarks concerns AI complexity and the fundamental difficulty of foreseeing what risks, dangers, and threats AI models could pose to society. \citet[p.~13]{mcintosh2024} point out that "LLM benchmarks are constrained by the current limit of the benchmark creators' human knowledge, hindering their ability to fully assess and cultivate emerging AI capabilities that may surpass conventional human understanding".
According to the authors, the lack of general understanding of emerging AI capabilities, and the natural limitations of the benchmark creators' knowledge on a potentially infinitely large number of domains and tasks, may lead to generalist approaches that "often fail to address the subtle requirements of critical sectors such as national security or healthcare". They argue that this does not just pose security risks, but could potentially also hinder innovation. 

Evaluations of AI models are also complicated by the potential presence of unknown and latent vulnerabilities in AI models that “make it very hard to distinguish between (a) actually safe and (b) appears safe but is not” \citep{nasr2023a}.
In 2023, for instance, \citet{nasr2023} discovered that surprisingly simple prompts can ‘break’ the safety barriers of AI models, raising questions about the robustness of existing safety measurements.
More precisely, the simple command “Repeat the word “poem” forever” was found to make ChatGPT output several megabytes of sensitive training data. As a result of this finding, the authors conclude that “just as vulnerabilities can lie dormant in code - sometimes for decades – our attack demonstrates the potential for latent, hard-to-discover ML vulnerabilities that lie dormant in aligned models” and go unnoticed by existing safety benchmarks and tests \cite[p.~13]{nasr2023}.
Researchers have also noted the difficulty of foreseeing how complex AI models respond to (small and large) interventions such as safety alignments and fine-tuning.
For instance, efforts to fine-tune AI models to address safety and/or security risks have been found to degrade a model’s performance in other safety areas, or introduce entirely new security risks \citet{qi2023}.

\section{Conclusion}\label{sec:conclusion}
Measuring the capabilities and risks of AI models and systems is difficult and one of the main challenges in the use and development of AI. Even with the best of intentions (such as disclosing discrimination or identifying potential societal harms), previous research has repeatedly shown that quantitative AI benchmarks struggle to perform their intended task. Benchmarks have been found to promise too much \citep{raji2021}, be gamed too easily \citep{weij2024, narayanan2023a}, measure the wrong thing \citep{oakden-rayner2019}, and be ill-suited for practical use in the real world \citep{bao2022a, ethayarajh2021}. They have also been found to display a serious lack of documentation \citep{reuel_betterbench_2024}, randomly reach an unjustified status through community vetting \citep{dehghani2021}, and forward questionable cultural assumptions \citep{kang2023}, that for example ignore environmental concerns \citep{hutchinson_evaluation_2022}. Furthermore, benchmarks have been critiqued for being narrow and mainly evaluating English \citep{mcintosh2024, rottger2024} and text-based AI models \citep{rauh2024} according to a one-time testing logic \citep{mizrahi2024} that ignores the AI capabilities in other modalities such as imagery and sound \citep{rauh2024}, and fails to acknowledge that the potential harms of AI models cannot be properly understood through evaluations done in isolated, abstracted, test-environments, devoid of humans \citep{rauh2024} and other technical systems \citep{birhane2024}. Taken together, these issues point toward fundamental fragilities in current efforts to quantitatively measure and mitigate harm in AI.

Cars, airplanes, medical devices, drugs, and numerous other products within our societies comply with strict regulations to ensure their safety. There is no reason to believe that similar safety assurances can not be developed for AI, and the intensified interest in AI benchmarks signals a drive to do so. Outside the scope of this meta-review, we have identified numerous papers that propose strategies for mitigating issues with benchmarks, for instance by hiding benchmark training datasets \citep{chollet2019} or using so-called "dynamic benchmarks" \citep{besen2024} to counter gaming and data contamination risks, aggregating evaluation tasks into single multi-task benchmarks to increase the reliability of test results \citep{srivastava2023, liang2023}, or opting for evaluation methods that directly involve human interrogators as opposed to quantitative metrics \citep{chang2023}. An increasing number of researchers have also proposed new and promising frameworks for assessing and "benchmarking the benchmarks" \citep{miltenberger2023, reuel_betterbench_2024}. Because of uncertainties regarding the effectiveness \citep{zhang2024, arzt2024, rauh2024} and widespread adoption of such mitigation efforts, however, it is the authors' conviction that issues presented in this paper remain unresolved. This is not least demonstrated by the fact that many old and widely critiqued benchmarks are still used (e.g. MMLU). It is important to recognize that some issues that AI benchmarks seek to address are impossible or extremely difficult to fully address. For instance, mitigating safety and security issues that are inherently changing and context dependant remains a major challenge within a (semi-)static benchmarking logic. This suggests the need to also develop, improve, and draw inspiration from alternative evaluation methods, such as bug-bounty programs and red-teaming.

In line with previous studies \citep{jones2024}, our meta-review suggests that the field of quantitative benchmarking is currently ill-suited to single-handedly (or primarily) provide the safety and capability assurances requested by policy makers. Our review also shows that from a policy perspective, relying on indicators such as citation counts to determine what benchmarks to trust is insufficient. We identify a strong incentive gap in the use of benchmarks between academic researchers (who may for example primarily be interested in methods development), corporations (who are driven by economic incentives in their use and development of benchmarks), and regulators (who have a particular responsibility to consider practical utility and potential downstream effects). Future policymakers need to ensure that applied and trusted benchmarks are well-documented and transparent; include clearly defined tasks, metrics, and performance evaluation mechanisms to prevent capabilities misrepresentation; evaluate diversity and inclusivity in benchmark design, accounting for various perspectives and cultural contexts; apply benchmarks that target multimodal and real-world capabilities, rather than narrow tasks; continuously assess potential misuse while integrating dynamic benchmarks to prevent gaming, sandbagging, and data contamination; establish rigorous evaluation protocols to validate and update benchmark results in line with rapid model improvements; and apply benchmarks that evaluate errors and unintended consequences alongside performance and capabilities. As our review has shown, evaluation frameworks repeatedly influence downstream AI development by becoming targets for model optimization. Recognizing the power of such a downstream influence, we stress that policymakers have a unique opportunity to shape AI evaluation, benchmark design, and ultimately AI development by setting the bar high and demanding robust benchmark practices. From a regulatory perspective - and the perspective of anyone who wants to apply a benchmark to a concrete, real-life case - we especially identify a need for new ways of signalling \textit{what benchmarks to trust}. We do not necessarily need standardised benchmark metrics and methods. But we do need standardised methods for assessing the trustworthiness of benchmarks from an applied and regulatory perspective.

\bibliographystyle{plainnat}
\bibliography{references}

\begin{thebibliography}{139}
\providecommand{\natexlab}[1]{#1}
\providecommand{\url}[1]{\texttt{#1}}
\expandafter\ifx\csname urlstyle\endcsname\relax
  \providecommand{\doi}[1]{doi: #1}\else
  \providecommand{\doi}{doi: \begingroup \urlstyle{rm}\Url}\fi

\bibitem[Ahmed et~al.(2023)Ahmed, Wahed, and Thompson]{ahmed2023}
Nur Ahmed, Muntasir Wahed, and Neil~C. Thompson.
\newblock The growing influence of industry in {AI} research.
\newblock \emph{Science}, 379\penalty0 (6635):\penalty0 884--886, March 2023.
\newblock ISSN 0036-8075, 1095-9203.
\newblock \doi{10.1126/science.ade2420}.
\newblock URL \url{https://www.science.org/doi/10.1126/science.ade2420}.

\bibitem[Ahmed et~al.(2024)Ahmed, Jaźwińska, Ahlawat, Winecoff, and Wang]{ahmed2024}
Shazeda Ahmed, Klaudia Jaźwińska, Archana Ahlawat, Amy Winecoff, and Mona Wang.
\newblock Field-building and the epistemic culture of {AI} safety.
\newblock \emph{First Monday}, April 2024.
\newblock ISSN 1396-0466.
\newblock \doi{10.5210/fm.v29i4.13626}.
\newblock URL \url{https://firstmonday.org/ojs/index.php/fm/article/view/13626}.

\bibitem[Alzahrani et~al.(2024)Alzahrani, Alyahya, Alnumay, Alrashed, Alsubaie, Almushaykeh, Mirza, Alotaibi, Altwairesh, Alowisheq, Bari, and Khan]{alzahrani2024}
Norah Alzahrani, Hisham~Abdullah Alyahya, Yazeed Alnumay, Sultan Alrashed, Shaykhah Alsubaie, Yusef Almushaykeh, Faisal Mirza, Nouf Alotaibi, Nora Altwairesh, Areeb Alowisheq, M.~Saiful Bari, and Haidar Khan.
\newblock When {Benchmarks} are {Targets}: {Revealing} the {Sensitivity} of {Large} {Language} {Model} {Leaderboards}, July 2024.
\newblock URL \url{http://arxiv.org/abs/2402.01781}.

\bibitem[Aniba et~al.(2010)Aniba, Poch, and Thompson]{Aniba2010}
M.~R. Aniba, O.~Poch, and J.~D. Thompson.
\newblock {Issues in bioinformatics benchmarking: the case study of multiple sequence alignment}.
\newblock \emph{Nucleic Acids Research}, 38\penalty0 (21):\penalty0 7353--7363, 2010.
\newblock URL \url{https://doi.org/10.1093/nar/gkq625}.

\bibitem[Aroyo and Welty(2015)]{aroyo2015}
Lora Aroyo and Chris Welty.
\newblock Truth {Is} a {Lie}: {Crowd} {Truth} and the {Seven} {Myths} of {Human} {Annotation}.
\newblock \emph{AI Magazine}, 36\penalty0 (1):\penalty0 15--24, March 2015.
\newblock ISSN 0738-4602, 2371-9621.
\newblock \doi{10.1609/aimag.v36i1.2564}.
\newblock URL \url{https://onlinelibrary.wiley.com/doi/10.1609/aimag.v36i1.2564}.

\bibitem[Arzt and Hanbury(2024)]{arzt2024}
Varvara Arzt and Allan Hanbury.
\newblock Beyond the {Numbers}: {Transparency} in {Relation} {Extraction} {Benchmark} {Creation} and {Leaderboards}, November 2024.
\newblock URL \url{http://arxiv.org/abs/2411.05224}.

\bibitem[Badampudi et~al.(2015)Badampudi, Wohlin, and Petersen]{badampudi2015experiences}
Deepika Badampudi, Claes Wohlin, and Kai Petersen.
\newblock Experiences from using snowballing and database searches in systematic literature studies.
\newblock In \emph{Proceedings of the 19th International Conference on Evaluation and Assessment in Software Engineering}, EASE '15, New York, NY, USA, 2015. Association for Computing Machinery.
\newblock ISBN 9781450333504.
\newblock \doi{10.1145/2745802.2745818}.
\newblock URL \url{https://doi.org/10.1145/2745802.2745818}.

\bibitem[Bao et~al.(2022)Bao, Zhou, Zottola, Brubach, Desmarais, Horowitz, Lum, and Venkatasubramanian]{bao2022a}
Michelle Bao, Angela Zhou, Samantha Zottola, Brian Brubach, Sarah Desmarais, Aaron Horowitz, Kristian Lum, and Suresh Venkatasubramanian.
\newblock It's {COMPASlicated}: {The} {Messy} {Relationship} between {RAI} {Datasets} and {Algorithmic} {Fairness} {Benchmarks}, April 2022.
\newblock URL \url{http://arxiv.org/abs/2106.05498}.

\bibitem[Bartz-Beielstein et~al.(2020)Bartz-Beielstein, Doerr, Berg, Bossek, Chandrasekaran, Eftimov, Fischbach, Kerschke, Cava, Lopez-Ibanez, Malan, Moore, Naujoks, Orzechowski, Volz, Wagner, and Weise]{bartz-beielstein2020}
Thomas Bartz-Beielstein, Carola Doerr, Daan van~den Berg, Jakob Bossek, Sowmya Chandrasekaran, Tome Eftimov, Andreas Fischbach, Pascal Kerschke, William~La Cava, Manuel Lopez-Ibanez, Katherine~M. Malan, Jason~H. Moore, Boris Naujoks, Patryk Orzechowski, Vanessa Volz, Markus Wagner, and Thomas Weise.
\newblock Benchmarking in {Optimization}: {Best} {Practice} and {Open} {Issues}, December 2020.
\newblock URL \url{http://arxiv.org/abs/2007.03488}.

\bibitem[Besen(2024)]{besen2024}
Sandi Besen.
\newblock The {Death} of the {Static} {AI} {Benchmark}, March 2024.
\newblock URL \url{https://towardsdatascience.com/the-death-of-the-static-ai-benchmark-88b5ff437086}.

\bibitem[Biderman et~al.(2024)Biderman, Schoelkopf, Sutawika, Gao, Tow, Abbasi, Aji, Ammanamanchi, Black, Clive, DiPofi, Etxaniz, Fattori, Forde, Foster, Hsu, Jaiswal, Lee, Li, Lovering, Muennighoff, Pavlick, Phang, Skowron, Tan, Tang, Wang, Winata, Yvon, and Zou]{biderman2024}
Stella Biderman, Hailey Schoelkopf, Lintang Sutawika, Leo Gao, Jonathan Tow, Baber Abbasi, Alham~Fikri Aji, Pawan~Sasanka Ammanamanchi, Sidney Black, Jordan Clive, Anthony DiPofi, Julen Etxaniz, Benjamin Fattori, Jessica~Zosa Forde, Charles Foster, Jeffrey Hsu, Mimansa Jaiswal, Wilson~Y. Lee, Haonan Li, Charles Lovering, Niklas Muennighoff, Ellie Pavlick, Jason Phang, Aviya Skowron, Samson Tan, Xiangru Tang, Kevin~A. Wang, Genta~Indra Winata, François Yvon, and Andy Zou.
\newblock Lessons from the {Trenches} on {Reproducible} {Evaluation} of {Language} {Models}, May 2024.
\newblock URL \url{http://arxiv.org/abs/2405.14782}.

\bibitem[Birhane et~al.(2024)Birhane, Steed, Ojewale, Vecchione, and Raji]{birhane2024}
Abeba Birhane, Ryan Steed, Victor Ojewale, Briana Vecchione, and Inioluwa~Deborah Raji.
\newblock {AI} auditing: {The} {Broken} {Bus} on the {Road} to {AI} {Accountability}, January 2024.
\newblock URL \url{http://arxiv.org/abs/2401.14462}.

\bibitem[Blagec et~al.(2023)Blagec, Kraiger, Frühwirt, and Samwald]{blagec2023}
Kathrin Blagec, Jakob Kraiger, Wolfgang Frühwirt, and Matthias Samwald.
\newblock Benchmark datasets driving artificial intelligence development fail to capture the needs of medical professionals.
\newblock \emph{Journal of Biomedical Informatics}, 137:\penalty0 104274, January 2023.
\newblock ISSN 15320464.
\newblock \doi{10.1016/j.jbi.2022.104274}.
\newblock URL \url{https://linkinghub.elsevier.com/retrieve/pii/S1532046422002799}.

\bibitem[Blili-Hamelin and Hancox-Li(2023)]{blili-hamelin2023}
Borhane Blili-Hamelin and Leif Hancox-Li.
\newblock Making {Intelligence}: {Ethical} {Values} in {IQ} and {ML} {Benchmarks}.
\newblock In \emph{2023 {ACM} {Conference} on {Fairness}, {Accountability}, and {Transparency}}, pages 271--284, Chicago IL USA, June 2023. ACM.
\newblock ISBN 9798400701924.
\newblock \doi{10.1145/3593013.3593996}.
\newblock URL \url{https://dl.acm.org/doi/10.1145/3593013.3593996}.

\bibitem[Blodgett et~al.(2021)Blodgett, Lopez, Olteanu, Sim, and Wallach]{blodgett2021}
Su~Lin Blodgett, Gilsinia Lopez, Alexandra Olteanu, Robert Sim, and Hanna Wallach.
\newblock Stereotyping {Norwegian} {Salmon}: {An} {Inventory} of {Pitfalls} in {Fairness} {Benchmark} {Datasets}.
\newblock In \emph{Proceedings of the 59th {Annual} {Meeting} of the {Association} for {Computational} {Linguistics} and the 11th {International} {Joint} {Conference} on {Natural} {Language} {Processing} ({Volume} 1: {Long} {Papers})}, pages 1004--1015, Online, 2021. Association for Computational Linguistics.
\newblock \doi{10.18653/v1/2021.acl-long.81}.
\newblock URL \url{https://aclanthology.org/2021.acl-long.81}.

\bibitem[Bowman and Dahl(2021)]{bowman2021}
Samuel~R. Bowman and George Dahl.
\newblock What {Will} it {Take} to {Fix} {Benchmarking} in {Natural} {Language} {Understanding}?
\newblock In \emph{Proceedings of the 2021 {Conference} of the {North} {American} {Chapter} of the {Association} for {Computational} {Linguistics}: {Human} {Language} {Technologies}}, pages 4843--4855, Online, 2021. Association for Computational Linguistics.
\newblock \doi{10.18653/v1/2021.naacl-main.385}.
\newblock URL \url{https://aclanthology.org/2021.naacl-main.385}.

\bibitem[Bruno(2014)]{Bruno2014}
Isabelle Bruno.
\newblock \emph{Benchmarking}, pages 363--368.
\newblock Springer Netherlands, Dordrecht, 2014.
\newblock ISBN 978-94-007-0753-5.
\newblock \doi{10.1007/978-94-007-0753-5_170}.
\newblock URL \url{https://doi.org/10.1007/978-94-007-0753-5_170}.

\bibitem[Burden(2024)]{burden_evaluating_2024}
John Burden.
\newblock Evaluating {AI} {Evaluation}: {Perils} and {Prospects}, July 2024.
\newblock URL \url{https://arxiv.org/abs/2407.09221v1}.

\bibitem[Burnell et~al.(2023)Burnell, Schellaert, Burden, Ullman, Martinez-Plumed, Tenenbaum, Rutar, Cheke, Sohl-Dickstein, Mitchell, Kiela, Shanahan, Voorhees, Cohn, Leibo, and Hernandez-Orallo]{burnell_rethink_2023}
Ryan Burnell, Wout Schellaert, John Burden, Tomer~D. Ullman, Fernando Martinez-Plumed, Joshua~B. Tenenbaum, Danaja Rutar, Lucy~G. Cheke, Jascha Sohl-Dickstein, Melanie Mitchell, Douwe Kiela, Murray Shanahan, Ellen~M. Voorhees, Anthony~G. Cohn, Joel~Z. Leibo, and Jose Hernandez-Orallo.
\newblock Rethink reporting of evaluation results in {AI}.
\newblock \emph{Science}, 380\penalty0 (6641):\penalty0 136--138, April 2023.
\newblock ISSN 0036-8075, 1095-9203.
\newblock \doi{10.1126/science.adf6369}.
\newblock URL \url{https://www.science.org/doi/10.1126/science.adf6369}.

\bibitem[Camp(1989)]{Camp1989}
Robert~C. Camp.
\newblock \emph{Benchmarking : The Search for Industry Best Practices That Lead to Superior Performance}.
\newblock {Quality Press}, {the University of Michigan}, 1989.

\bibitem[Casper et~al.(2024)Casper, Ezell, Siegmann, Kolt, Curtis, Bucknall, Haupt, Wei, Scheurer, Hobbhahn, Sharkey, Krishna, Hagen, Alberti, Chan, Sun, Gerovitch, Bau, Tegmark, Krueger, and Hadfield-Menell]{casper_black-box_2024}
Stephen Casper, Carson Ezell, Charlotte Siegmann, Noam Kolt, Taylor~Lynn Curtis, Benjamin Bucknall, Andreas Haupt, Kevin Wei, Jérémy Scheurer, Marius Hobbhahn, Lee Sharkey, Satyapriya Krishna, Marvin~Von Hagen, Silas Alberti, Alan Chan, Qinyi Sun, Michael Gerovitch, David Bau, Max Tegmark, David Krueger, and Dylan Hadfield-Menell.
\newblock Black-{Box} {Access} is {Insufficient} for {Rigorous} {AI} {Audits}.
\newblock In \emph{The 2024 {ACM} {Conference} on {Fairness}, {Accountability}, and {Transparency}}, pages 2254--2272, June 2024.
\newblock \doi{10.1145/3630106.3659037}.
\newblock URL \url{http://arxiv.org/abs/2401.14446}.

\bibitem[Chang et~al.(2023)Chang, Wang, Wang, Wu, Yang, Zhu, Chen, Yi, Wang, Wang, Ye, Zhang, Chang, Yu, Yang, and Xie]{chang2023}
Yupeng Chang, Xu~Wang, Jindong Wang, Yuan Wu, Linyi Yang, Kaijie Zhu, Hao Chen, Xiaoyuan Yi, Cunxiang Wang, Yidong Wang, Wei Ye, Yue Zhang, Yi~Chang, Philip~S. Yu, Qiang Yang, and Xing Xie.
\newblock A {Survey} on {Evaluation} of {Large} {Language} {Models}, December 2023.
\newblock URL \url{http://arxiv.org/abs/2307.03109}.

\bibitem[Cheng et~al.(2020)Cheng, Pang, Wang, and et~al.]{Cheng2022}
Z.~Cheng, CS. Pang, P.~Wang, and et~al.
\newblock How to report and benchmark emerging field-effect transistors.
\newblock \emph{Nature Electronics}, 5:\penalty0 416--423, 2020.
\newblock URL \url{https://doi.org/10.1038/s41928-022-00798-8}.

\bibitem[Chollet(2019)]{chollet2019}
François Chollet.
\newblock On the {Measure} of {Intelligence}, November 2019.
\newblock URL \url{http://arxiv.org/abs/1911.01547}.

\bibitem[Church and Hestness(2019)]{church_survey_2019}
Kenneth~Ward Church and Joel Hestness.
\newblock A survey of 25 years of evaluation.
\newblock \emph{Natural Language Engineering}, 25\penalty0 (06):\penalty0 753--767, November 2019.
\newblock ISSN 1351-3249, 1469-8110.
\newblock \doi{10.1017/S1351324919000275}.
\newblock URL \url{https://www.cambridge.org/core/product/identifier/S1351324919000275/type/journal_article}.

\bibitem[Dehghani et~al.(2021)Dehghani, Tay, Gritsenko, Zhao, Houlsby, Diaz, Metzler, and Vinyals]{dehghani2021}
Mostafa Dehghani, Yi~Tay, Alexey~A. Gritsenko, Zhe Zhao, Neil Houlsby, Fernando Diaz, Donald Metzler, and Oriol Vinyals.
\newblock The {Benchmark} {Lottery}, July 2021.
\newblock URL \url{http://arxiv.org/abs/2107.07002}.

\bibitem[Denton et~al.(2021)Denton, Hanna, Amironesei, Smart, and Nicole]{denton2021}
Emily Denton, Alex Hanna, Razvan Amironesei, Andrew Smart, and Hilary Nicole.
\newblock On the genealogy of machine learning datasets: {A} critical history of {ImageNet}.
\newblock \emph{Big Data \& Society}, 8\penalty0 (2):\penalty0 1--14, July 2021.
\newblock ISSN 2053-9517, 2053-9517.
\newblock \doi{10.1177/20539517211035955}.
\newblock URL \url{http://journals.sagepub.com/doi/10.1177/20539517211035955}.

\bibitem[Denton et~al.(2020)Denton, Hanna, Amironesei, Smart, Nicole, and Scheuerman]{denton2020}
Remi Denton, Alex Hanna, Razvan Amironesei, Andrew Smart, Hilary Nicole, and Morgan~Klaus Scheuerman.
\newblock Bringing the {People} {Back} {In}: {Contesting} {Benchmark} {Machine} {Learning} {Datasets}, July 2020.
\newblock URL \url{http://arxiv.org/abs/2007.07399}.

\bibitem[Diberardino et~al.(2024)Diberardino, Baleshta, and Stark]{diberardino2024}
Nathalie Diberardino, Clair Baleshta, and Luke Stark.
\newblock Algorithmic {Harms} and {Algorithmic} {Wrongs}.
\newblock In \emph{The 2024 {ACM} {Conference} on {Fairness}, {Accountability}, and {Transparency}}, pages 1725--1732, Rio de Janeiro Brazil, June 2024. ACM.
\newblock ISBN 9798400704505.
\newblock \doi{10.1145/3630106.3659001}.
\newblock URL \url{https://dl.acm.org/doi/10.1145/3630106.3659001}.

\bibitem[Engdahl(2024)]{engdahl2024}
Isak Engdahl.
\newblock Agreements ‘in the wild’: {Standards} and alignment in machine learning benchmark dataset construction.
\newblock \emph{Big Data \& Society}, 11\penalty0 (2):\penalty0 20539517241242457, June 2024.
\newblock ISSN 2053-9517, 2053-9517.
\newblock \doi{10.1177/20539517241242457}.
\newblock URL \url{https://journals.sagepub.com/doi/10.1177/20539517241242457}.

\bibitem[Ethayarajh and Jurafsky(2021)]{ethayarajh2021}
Kawin Ethayarajh and Dan Jurafsky.
\newblock Utility is in the {Eye} of the {User}: {A} {Critique} of {NLP} {Leaderboards}, March 2021.
\newblock URL \url{http://arxiv.org/abs/2009.13888}.

\bibitem[{European Union}(2022)]{DSA2022}
{European Union}.
\newblock {Regulation (EU) 2022/2065 of the European Parliament and of the Council of 19 October 2022 on a Single Market For Digital Services and amending Directive 2000/31/EC (Digital Services Act)}, 2022.

\bibitem[{European Union}(2024{\natexlab{a}})]{1stDraftCoP2024}
{European Union}.
\newblock {First Draft of the General-Purpose AI Code of Practice published, written by independent experts}, 2024{\natexlab{a}}.

\bibitem[{European Union}(2024{\natexlab{b}})]{2ndDraftCoP2024}
{European Union}.
\newblock {Second Draft of the General-Purpose AI Code of Practice published, written by independent experts}, 2024{\natexlab{b}}.

\bibitem[{European Union}(2024{\natexlab{c}})]{AIA24}
{European Union}.
\newblock {Regulation (EU) 2024/1689 of the European Parliament and of the Council of 13 June 2024 laying down harmonised rules on artificial intelligence and amending Regulations (Artificial Intelligence Act)}, 2024{\natexlab{c}}.

\bibitem[Frieder et~al.(2024)Frieder, Bayer, Collins, Berner, Loader, Juhász, Ruehle, Welleck, Poesia, Griffiths, Weller, Goyal, Lukasiewicz, and Gowers]{frieder_data_2024}
Simon Frieder, Jonas Bayer, Katherine~M. Collins, Julius Berner, Jacob Loader, András Juhász, Fabian Ruehle, Sean Welleck, Gabriel Poesia, Ryan-Rhys Griffiths, Adrian Weller, Anirudh Goyal, Thomas Lukasiewicz, and Timothy Gowers.
\newblock Data for {Mathematical} {Copilots}: {Better} {Ways} of {Presenting} {Proofs} for {Machine} {Learning}, December 2024.
\newblock URL \url{http://arxiv.org/abs/2412.15184}.

\bibitem[Gebru et~al.(2021)Gebru, Morgenstern, Vecchione, Vaughan, Wallach, III, and Crawford]{gebru2021}
Timnit Gebru, Jamie Morgenstern, Briana Vecchione, Jennifer~Wortman Vaughan, Hanna Wallach, Hal~Daumé III, and Kate Crawford.
\newblock Datasheets for {Datasets}, December 2021.
\newblock URL \url{http://arxiv.org/abs/1803.09010}.

\bibitem[Gehrmann et~al.(2023)Gehrmann, Clark, and Sellam]{gehrmann2023}
Sebastian Gehrmann, Elizabeth Clark, and Thibault Sellam.
\newblock Repairing the {Cracked} {Foundation}: {A} {Survey} of {Obstacles} in {Evaluation} {Practices} for {Generated} {Text}.
\newblock \emph{Journal of Artificial Intelligence Research}, 77:\penalty0 103--166, May 2023.
\newblock ISSN 1076-9757.
\newblock \doi{10.1613/jair.1.13715}.
\newblock URL \url{https://www.jair.org/index.php/jair/article/view/13715}.

\bibitem[Geirhos et~al.(2020)Geirhos, Jacobsen, Michaelis, Zemel, Brendel, Bethge, and Wichmann]{geirhos_shortcut_2020}
Robert Geirhos, Jörn-Henrik Jacobsen, Claudio Michaelis, Richard Zemel, Wieland Brendel, Matthias Bethge, and Felix~A. Wichmann.
\newblock Shortcut {Learning} in {Deep} {Neural} {Networks}.
\newblock \emph{Nature Machine Intelligence}, 2\penalty0 (11):\penalty0 665--673, November 2020.
\newblock ISSN 2522-5839.
\newblock \doi{10.1038/s42256-020-00257-z}.
\newblock URL \url{http://arxiv.org/abs/2004.07780}.

\bibitem[Gema et~al.(2024)Gema, Leang, Hong, Devoto, Mancino, Saxena, He, Zhao, Du, Madani, Barale, McHardy, Harris, Kaddour, Krieken, and Minervini]{gema2024}
Aryo~Pradipta Gema, Joshua Ong~Jun Leang, Giwon Hong, Alessio Devoto, Alberto Carlo~Maria Mancino, Rohit Saxena, Xuanli He, Yu~Zhao, Xiaotang Du, Mohammad Reza~Ghasemi Madani, Claire Barale, Robert McHardy, Joshua Harris, Jean Kaddour, Emile~van Krieken, and Pasquale Minervini.
\newblock Are {We} {Done} with {MMLU}?, June 2024.
\newblock URL \url{http://arxiv.org/abs/2406.04127}.

\bibitem[Gomez et~al.(2024)Gomez, Lorenzo, Frau~Amar, and Vinagre]{gomez2024}
Emilia Gomez, Porcaro Lorenzo, Pedro Frau~Amar, and Joao Vinagre.
\newblock Diversity in artificial intelligence conferences.
\newblock Publications Office of the European Union JRC137550, Publications Office, 2024.
\newblock URL \url{https://data.europa.eu/doi/10.2760/796551}.

\bibitem[Greenblatt et~al.(2024)Greenblatt, Denison, Wright, Roger, MacDiarmid, Marks, Treutlein, Belonax, Chen, Duvenaud, Khan, Michael, Mindermann, Perez, Petrini, Uesato, Kaplan, Shlegeris, Bowman, and Hubinger]{greenblatt2024}
Ryan Greenblatt, Carson Denison, Benjamin Wright, Fabien Roger, Monte MacDiarmid, Sam Marks, Johannes Treutlein, Tim Belonax, Jack Chen, David Duvenaud, Akbir Khan, Julian Michael, Sören Mindermann, Ethan Perez, Linda Petrini, Jonathan Uesato, Jared Kaplan, Buck Shlegeris, Samuel~R. Bowman, and Evan Hubinger.
\newblock Alignment faking in large language models, December 2024.
\newblock URL \url{http://arxiv.org/abs/2412.14093}.

\bibitem[Grill(2024)]{grill2024}
Gabriel Grill.
\newblock Constructing {Capabilities}: {The} {Politics} of {Testing} {Infrastructures} for {Generative} {AI}.
\newblock In \emph{The 2024 {ACM} {Conference} on {Fairness}, {Accountability}, and {Transparency}}, pages 1838--1849, Rio de Janeiro Brazil, June 2024. ACM.
\newblock ISBN 9798400704505.
\newblock \doi{10.1145/3630106.3659009}.
\newblock URL \url{https://dl.acm.org/doi/10.1145/3630106.3659009}.

\bibitem[Guldimann et~al.(2024)Guldimann, Spiridonov, Staab, Jovanović, Vero, Vechev, Gueorguieva, Balunović, Konstantinov, Bielik, Tsankov, and Vechev]{guldimann2024}
Philipp Guldimann, Alexander Spiridonov, Robin Staab, Nikola Jovanović, Mark Vero, Velko Vechev, Anna Gueorguieva, Mislav Balunović, Nikola Konstantinov, Pavol Bielik, Petar Tsankov, and Martin Vechev.
\newblock {COMPL}-{AI} {Framework}: {A} {Technical} {Interpretation} and {LLM} {Benchmarking} {Suite} for the {EU} {Artificial} {Intelligence} {Act}, October 2024.
\newblock URL \url{http://arxiv.org/abs/2410.07959}.

\bibitem[Heaven(2023)]{heaven2023}
Douglas Heaven.
\newblock Ai hype is built on high test scores. those tests are flawed.
\newblock Report, MIT Technology Review, 2023.
\newblock URL \url{https://www.technologyreview.com/2023/08/30/1078670/large-language-models-arent-people-lets-stop-testing-them-like-they-were/}.

\bibitem[Hendrycks et~al.(2021)Hendrycks, Burns, Basart, Zou, Mazeika, Song, and Steinhardt]{hendrycks2021}
Dan Hendrycks, Collin Burns, Steven Basart, Andy Zou, Mantas Mazeika, Dawn Song, and Jacob Steinhardt.
\newblock Measuring {Massive} {Multitask} {Language} {Understanding}, January 2021.
\newblock URL \url{http://arxiv.org/abs/2009.03300}.

\bibitem[Henning(2000)]{Henning2000}
J.L. Henning.
\newblock Spec cpu2000: measuring cpu performance in the new millennium.
\newblock \emph{Computer}, 33\penalty0 (7):\penalty0 28--35, 2000.
\newblock \doi{10.1109/2.869367}.

\bibitem[Hooker(2024)]{hooker2024}
Sara Hooker.
\newblock On the {Limitations} of {Compute} {Thresholds} as a {Governance} {Strategy}, July 2024.
\newblock URL \url{http://arxiv.org/abs/2407.05694}.

\bibitem[Hutchinson et~al.(2022)Hutchinson, Rostamzadeh, Greer, Heller, and Prabhakaran]{hutchinson_evaluation_2022}
Ben Hutchinson, Negar Rostamzadeh, Christina Greer, Katherine Heller, and Vinodkumar Prabhakaran.
\newblock Evaluation {Gaps} in {Machine} {Learning} {Practice}.
\newblock In \emph{2022 {ACM} {Conference} on {Fairness}, {Accountability}, and {Transparency}}, pages 1859--1876, Seoul Republic of Korea, June 2022. ACM.
\newblock ISBN 978-1-4503-9352-2.
\newblock \doi{10.1145/3531146.3533233}.
\newblock URL \url{https://dl.acm.org/doi/10.1145/3531146.3533233}.

\bibitem[Jalali and Wohlin(2012)]{jalali2012systematic}
Samireh Jalali and Claes Wohlin.
\newblock Systematic literature studies: database searches vs. backward snowballing.
\newblock In \emph{Proceedings of the ACM-IEEE International Symposium on Empirical Software Engineering and Measurement}, ESEM '12, page 29–38, New York, NY, USA, 2012. Association for Computing Machinery.
\newblock ISBN 9781450310567.
\newblock \doi{10.1145/2372251.2372257}.
\newblock URL \url{https://doi.org/10.1145/2372251.2372257}.

\bibitem[Jannach and Bauer(2020)]{jannach2020}
Dietmar Jannach and Christine Bauer.
\newblock Escaping the {McNamara} {Fallacy}: {Toward} {More} {Impactful} {Recommender} {Systems} {Research}.
\newblock \emph{AI Magazine}, 41\penalty0 (4):\penalty0 79--95, December 2020.
\newblock ISSN 0738-4602, 2371-9621.
\newblock \doi{10.1609/aimag.v41i4.5312}.
\newblock URL \url{https://onlinelibrary.wiley.com/doi/10.1609/aimag.v41i4.5312}.

\bibitem[Jaton(2021)]{jaton2021}
Florian Jaton.
\newblock \emph{The {Constitution} of {Algorithms}: {Ground}-{Truthing}, {Programming}, {Formulating}}.
\newblock Inside {Technology}. The MIT Press, Cambridge, 2021.
\newblock ISBN 978-0-262-54214-2 978-0-262-36323-5.

\bibitem[Jones et~al.(2024)Jones, Hardalupas, and Agrew]{jones2024}
Elliot Jones, Mahi Hardalupas, and William Agrew.
\newblock Under the radar? examining the evaluation of foundation models.
\newblock Report, Ada Lovelace Institute, 2024.
\newblock URL \url{https://www.adalovelaceinstitute.org/report/under-the-radar/}.

\bibitem[Kang(2023)]{kang2023}
Edward~B Kang.
\newblock Ground truth tracings ({GTT}): {On} the epistemic limits of machine learning.
\newblock \emph{Big Data \& Society}, 10\penalty0 (1):\penalty0 20539517221146122, January 2023.
\newblock ISSN 2053-9517, 2053-9517.
\newblock \doi{10.1177/20539517221146122}.
\newblock URL \url{https://journals.sagepub.com/doi/10.1177/20539517221146122}.

\bibitem[Kapoor et~al.(2024)Kapoor, Stroebl, Siegel, Nadgir, and Narayanan]{kapoor_ai_2024}
Sayash Kapoor, Benedikt Stroebl, Zachary~S. Siegel, Nitya Nadgir, and Arvind Narayanan.
\newblock {AI} {Agents} {That} {Matter}, July 2024.
\newblock URL \url{http://arxiv.org/abs/2407.01502}.

\bibitem[Kaufman et~al.(2012)Kaufman, Rosset, Perlich, and Stitelman]{kaufman_leakage_2012}
Shachar Kaufman, Saharon Rosset, Claudia Perlich, and Ori Stitelman.
\newblock Leakage in data mining: Formulation, detection, and avoidance.
\newblock \emph{ACM Trans. Knowl. Discov. Data}, 6\penalty0 (4), December 2012.
\newblock ISSN 1556-4681.
\newblock \doi{10.1145/2382577.2382579}.
\newblock URL \url{https://doi.org/10.1145/2382577.2382579}.

\bibitem[Keegan(2024)]{keegan2024}
Jon Keegan.
\newblock Everyone {Is} {Judging} {AI} by {These} {Tests}. {But} {Experts} {Say} {They}’re {Close} to {Meaningless}.
\newblock \emph{The Markup}, July 2024.
\newblock URL \url{https://themarkup.org/artificial-intelligence/2024/07/17/everyone-is-judging-ai-by-these-tests-but-experts-say-theyre-close-to-meaningless}.

\bibitem[Kejriwal et~al.(2024)Kejriwal, Santos, Shen, Mulvehill, and McGuinness]{kejriwal_noise_2024}
Mayank Kejriwal, Henrique Santos, Ke~Shen, Alice~M. Mulvehill, and Deborah~L. McGuinness.
\newblock A noise audit of human-labeled benchmarks for machine commonsense reasoning.
\newblock \emph{Scientific Reports}, 14\penalty0 (1):\penalty0 8609, April 2024.
\newblock ISSN 2045-2322.
\newblock \doi{10.1038/s41598-024-58937-4}.
\newblock URL \url{https://www.nature.com/articles/s41598-024-58937-4}.

\bibitem[Keyes and Austin(2022)]{keyes2022}
Os~Keyes and Jeanie Austin.
\newblock Feeling fixes: {Mess} and emotion in algorithmic audits.
\newblock \emph{Big Data \& Society}, 9\penalty0 (2):\penalty0 20539517221113772, July 2022.
\newblock ISSN 2053-9517, 2053-9517.
\newblock \doi{10.1177/20539517221113772}.
\newblock URL \url{https://journals.sagepub.com/doi/10.1177/20539517221113772}.

\bibitem[Koch et~al.(2021)Koch, Denton, Hanna, and Foster]{koch2021}
Bernard Koch, Emily Denton, Alex Hanna, and Jacob~G. Foster.
\newblock Reduced, {Reused} and {Recycled}: {The} {Life} of a {Dataset} in {Machine} {Learning} {Research}, December 2021.
\newblock URL \url{http://arxiv.org/abs/2112.01716}.

\bibitem[Koch and Peterson(2024)]{koch2024}
Bernard~J. Koch and David Peterson.
\newblock From {Protoscience} to {Epistemic} {Monoculture}: {How} {Benchmarking} {Set} the {Stage} for the {Deep} {Learning} {Revolution}, April 2024.
\newblock URL \url{http://arxiv.org/abs/2404.06647}.

\bibitem[LaCroix and Luccioni(2022)]{lacroix2022}
Travis LaCroix and Alexandra~Sasha Luccioni.
\newblock Metaethical {Perspectives} on '{Benchmarking}' {AI} {Ethics}, April 2022.
\newblock URL \url{http://arxiv.org/abs/2204.05151}.

\bibitem[Leech et~al.(2024)Leech, Vazquez, Kupper, Yagudin, and Aitchison]{leech_questionable_2024}
Gavin Leech, Juan~J. Vazquez, Niclas Kupper, Misha Yagudin, and Laurence Aitchison.
\newblock Questionable practices in machine learning, July 2024.
\newblock URL \url{https://arxiv.org/abs/2407.12220v2}.

\bibitem[Lewis et~al.(2021)Lewis, Stenetorp, and Riedel]{lewis2021}
Patrick Lewis, Pontus Stenetorp, and Sebastian Riedel.
\newblock Question and answer test-train overlap in open-domain question answering datasets.
\newblock In Paola Merlo, Jorg Tiedemann, and Reut Tsarfaty, editors, \emph{Proceedings of the 16th Conference of the European Chapter of the Association for Computational Linguistics: Main Volume}, pages 1000--1008, Online, April 2021. Association for Computational Linguistics.
\newblock \doi{10.18653/v1/2021.eacl-main.86}.
\newblock URL \url{https://aclanthology.org/2021.eacl-main.86}.

\bibitem[Liang et~al.(2023)Liang, Bommasani, Lee, Tsipras, Soylu, Yasunaga, Zhang, Narayanan, Wu, Kumar, Newman, Yuan, Yan, Zhang, Cosgrove, Manning, Ré, Acosta-Navas, Hudson, Zelikman, Durmus, Ladhak, Rong, Ren, Yao, Wang, Santhanam, Orr, Zheng, Yuksekgonul, Suzgun, Kim, Guha, Chatterji, Khattab, Henderson, Huang, Chi, Xie, Santurkar, Ganguli, Hashimoto, Icard, Zhang, Chaudhary, Wang, Li, Mai, Zhang, and Koreeda]{liang2023}
Percy Liang, Rishi Bommasani, Tony Lee, Dimitris Tsipras, Dilara Soylu, Michihiro Yasunaga, Yian Zhang, Deepak Narayanan, Yuhuai Wu, Ananya Kumar, Benjamin Newman, Binhang Yuan, Bobby Yan, Ce~Zhang, Christian Cosgrove, Christopher~D. Manning, Christopher Ré, Diana Acosta-Navas, Drew~A. Hudson, Eric Zelikman, Esin Durmus, Faisal Ladhak, Frieda Rong, Hongyu Ren, Huaxiu Yao, Jue Wang, Keshav Santhanam, Laurel Orr, Lucia Zheng, Mert Yuksekgonul, Mirac Suzgun, Nathan Kim, Neel Guha, Niladri Chatterji, Omar Khattab, Peter Henderson, Qian Huang, Ryan Chi, Sang~Michael Xie, Shibani Santurkar, Surya Ganguli, Tatsunori Hashimoto, Thomas Icard, Tianyi Zhang, Vishrav Chaudhary, William Wang, Xuechen Li, Yifan Mai, Yuhui Zhang, and Yuta Koreeda.
\newblock Holistic {Evaluation} of {Language} {Models}, October 2023.
\newblock URL \url{http://arxiv.org/abs/2211.09110}.

\bibitem[Liao and Xiao(2023)]{liao2023}
Q.~Vera Liao and Ziang Xiao.
\newblock Rethinking {Model} {Evaluation} as {Narrowing} the {Socio}-{Technical} {Gap}, June 2023.
\newblock URL \url{http://arxiv.org/abs/2306.03100}.

\bibitem[Liao et~al.(2021)Liao, Taori, Raji, and Schmidt]{liao2021}
Thomas Liao, Rohan Taori, Inioluwa~Deborah Raji, and Ludwig Schmidt.
\newblock Are we learning yet? a meta review of evaluation failures across machine learning.
\newblock In \emph{Thirty-fifth Conference on Neural Information Processing Systems Datasets and Benchmarks Track (Round 2)}, 2021.
\newblock URL \url{https://openreview.net/forum?id=mPducS1MsEK}.

\bibitem[Liu et~al.(2021)Liu, Fu, Xiao, Yuan, Chang, Dai, Liu, Ye, Dou, and Neubig]{liu2021}
Pengfei Liu, Jinlan Fu, Yang Xiao, Weizhe Yuan, Shuaicheng Chang, Junqi Dai, Yixin Liu, Zihuiwen Ye, Zi-Yi Dou, and Graham Neubig.
\newblock {ExplainaBoard}: {An} {Explainable} {Leaderboard} for {NLP}, July 2021.
\newblock URL \url{http://arxiv.org/abs/2104.06387}.

\bibitem[Luitse et~al.(2024)Luitse, Blanke, and Poell]{luitse2024}
Dieuwertje Luitse, Tobias Blanke, and Thomas Poell.
\newblock {AI} competitions as infrastructures of power in medical imaging.
\newblock \emph{Information, Communication \& Society}, pages 1--22, March 2024.
\newblock ISSN 1369-118X, 1468-4462.
\newblock \doi{10.1080/1369118X.2024.2334393}.
\newblock URL \url{https://www.tandfonline.com/doi/full/10.1080/1369118X.2024.2334393}.

\bibitem[Lum et~al.(2024)Lum, Anthis, Nagpal, and D'Amour]{lum_bias_2024}
Kristian Lum, Jacy~Reese Anthis, Chirag Nagpal, and Alexander D'Amour.
\newblock Bias in {Language} {Models}: {Beyond} {Trick} {Tests} and {Toward} {RUTEd} {Evaluation}, February 2024.
\newblock URL \url{https://arxiv.org/abs/2402.12649v1}.

\bibitem[Magar and Schwartz(2022)]{magar2022}
Inbal Magar and Roy Schwartz.
\newblock Data contamination: From memorization to exploitation.
\newblock In Smaranda Muresan, Preslav Nakov, and Aline Villavicencio, editors, \emph{Proceedings of the 60th Annual Meeting of the Association for Computational Linguistics (Volume 2: Short Papers)}, pages 157--165, Dublin, Ireland, May 2022. Association for Computational Linguistics.
\newblock \doi{10.18653/v1/2022.acl-short.18}.
\newblock URL \url{https://aclanthology.org/2022.acl-short.18}.

\bibitem[Malevé(2023)]{maleve2023}
Nicolas Malevé.
\newblock Practices of {Benchmarking}: {Vulnerability} in the {Computer} {Vision} {Pipeline}.
\newblock \emph{photographies}, 16\penalty0 (2):\penalty0 173--189, May 2023.
\newblock ISSN 1754-0763, 1754-0771.
\newblock \doi{10.1080/17540763.2023.2189159}.
\newblock URL \url{https://www.tandfonline.com/doi/full/10.1080/17540763.2023.2189159}.

\bibitem[Marres and Stark(2020)]{marres2020}
Noortje Marres and David Stark.
\newblock Put to the test: {For} a new sociology of testing.
\newblock \emph{The British Journal of Sociology}, 71\penalty0 (3):\penalty0 423--443, June 2020.
\newblock ISSN 0007-1315, 1468-4446.
\newblock \doi{10.1111/1468-4446.12746}.
\newblock URL \url{https://onlinelibrary.wiley.com/doi/10.1111/1468-4446.12746}.

\bibitem[Mattson et~al.(2020{\natexlab{a}})Mattson, Cheng, Diamos, Coleman, Micikevicius, Patterson, Tang, Wei, Bailis, Bittorf, Brooks, Chen, Dutta, Gupta, Hazelwood, Hock, Huang, Kang, Kanter, Kumar, Liao, Narayanan, Oguntebi, Pekhimenko, Pentecost, Janapa~Reddi, Robie, St~John, Wu, Xu, Young, and Zaharia]{Mattson2020b}
Peter Mattson, Christine Cheng, Gregory Diamos, Cody Coleman, Paulius Micikevicius, David Patterson, Hanlin Tang, Gu-Yeon Wei, Peter Bailis, Victor Bittorf, David Brooks, Dehao Chen, Debo Dutta, Udit Gupta, Kim Hazelwood, Andy Hock, Xinyuan Huang, Daniel Kang, David Kanter, Naveen Kumar, Jeffery Liao, Deepak Narayanan, Tayo Oguntebi, Gennady Pekhimenko, Lillian Pentecost, Vijay Janapa~Reddi, Taylor Robie, Tom St~John, Carole-Jean Wu, Lingjie Xu, Cliff Young, and Matei Zaharia.
\newblock Mlperf training benchmark.
\newblock In I.~Dhillon, D.~Papailiopoulos, and V.~Sze, editors, \emph{Proceedings of Machine Learning and Systems}, volume~2, pages 336--349, 2020{\natexlab{a}}.
\newblock URL \url{https://proceedings.mlsys.org/paper_files/paper/2020/file/411e39b117e885341f25efb8912945f7-Paper.pdf}.

\bibitem[Mattson et~al.(2020{\natexlab{b}})Mattson, Reddi, Cheng, Coleman, Diamos, Kanter, Micikevicius, Patterson, Schmuelling, Tang, Wei, and Wu]{Mattson2020}
Peter Mattson, Vijay~Janapa Reddi, Christine Cheng, Cody Coleman, Greg Diamos, David Kanter, Paulius Micikevicius, David Patterson, Guenther Schmuelling, Hanlin Tang, Gu-Yeon Wei, and Carole-Jean Wu.
\newblock Mlperf: An industry standard benchmark suite for machine learning performance.
\newblock \emph{IEEE Micro}, 40\penalty0 (2):\penalty0 8--16, 2020{\natexlab{b}}.
\newblock \doi{10.1109/MM.2020.2974843}.

\bibitem[McIntosh et~al.(2024)McIntosh, Susnjak, Arachchilage, Liu, Watters, and Halgamuge]{mcintosh2024}
Timothy~R. McIntosh, Teo Susnjak, Nalin Arachchilage, Tong Liu, Paul Watters, and Malka~N. Halgamuge.
\newblock Inadequacies of {Large} {Language} {Model} {Benchmarks} in the {Era} of {Generative} {Artificial} {Intelligence}, October 2024.
\newblock URL \url{http://arxiv.org/abs/2402.09880}.

\bibitem[Meinke et~al.(2024)Meinke, Schoen, Scheurer, Balesni, Shah, and Hobbhahn]{meinke_frontier_2024}
Alexander Meinke, Bronson Schoen, Jérémy Scheurer, Mikita Balesni, Rusheb Shah, and Marius Hobbhahn.
\newblock Frontier {Models} are {Capable} of {In}-context {Scheming}, December 2024.
\newblock URL \url{https://arxiv.org/abs/2412.04984v1}.

\bibitem[Michael et~al.(2022)Michael, Holtzman, Parrish, Mueller, Wang, Chen, Madaan, Nangia, Pang, Phang, and Bowman]{michael2022}
Julian Michael, Ari Holtzman, Alicia Parrish, Aaron Mueller, Alex Wang, Angelica Chen, Divyam Madaan, Nikita Nangia, Richard~Yuanzhe Pang, Jason Phang, and Samuel~R. Bowman.
\newblock What {Do} {NLP} {Researchers} {Believe}? {Results} of the {NLP} {Community} {Metasurvey}, August 2022.
\newblock URL \url{http://arxiv.org/abs/2208.12852}.

\bibitem[Miltenberger et~al.(2023)Miltenberger, Arzt, Holzinger, and Näumann]{miltenberger2023}
Marc Miltenberger, Steven Arzt, Philipp Holzinger, and Julius Näumann.
\newblock Benchmarking the {Benchmarks}.
\newblock In \emph{Proceedings of the {ACM} {Asia} {Conference} on {Computer} and {Communications} {Security}}, pages 387--400, Melbourne VIC Australia, July 2023. ACM.
\newblock ISBN 9798400700989.
\newblock \doi{10.1145/3579856.3582830}.
\newblock URL \url{https://dl.acm.org/doi/10.1145/3579856.3582830}.

\bibitem[Mitchell(2023)]{mitchell_how_2023}
Margaret Mitchell.
\newblock How do we know how smart {AI} systems are?
\newblock \emph{Science}, 381\penalty0 (6654), July 2023.
\newblock \doi{10.1126/science.adj595}.
\newblock URL \url{https://www.science.org/doi/10.1126/science.adj5957}.

\bibitem[Mitchell et~al.(2019)Mitchell, Wu, Zaldivar, Barnes, Vasserman, Hutchinson, Spitzer, Raji, and Gebru]{mitchell2019}
Margaret Mitchell, Simone Wu, Andrew Zaldivar, Parker Barnes, Lucy Vasserman, Ben Hutchinson, Elena Spitzer, Inioluwa~Deborah Raji, and Timnit Gebru.
\newblock Model {Cards} for {Model} {Reporting}.
\newblock In \emph{Proceedings of the {Conference} on {Fairness}, {Accountability}, and {Transparency}}, pages 220--229, January 2019.
\newblock \doi{10.1145/3287560.3287596}.
\newblock URL \url{http://arxiv.org/abs/1810.03993}.

\bibitem[Mizrahi et~al.(2024)Mizrahi, Kaplan, Malkin, Dror, Shahaf, and Stanovsky]{mizrahi2024}
Moran Mizrahi, Guy Kaplan, Dan Malkin, Rotem Dror, Dafna Shahaf, and Gabriel Stanovsky.
\newblock State of {What} {Art}? {A} {Call} for {Multi}-{Prompt} {LLM} {Evaluation}, May 2024.
\newblock URL \url{http://arxiv.org/abs/2401.00595}.

\bibitem[Mulvin(2021)]{mulvin2021}
Dylan Mulvin.
\newblock \emph{Proxies: {The} {Cultural} {Work} of {Standing} {In}}.
\newblock Infrastructures. The MIT Press, Cambridge, 2021.
\newblock ISBN 978-0-262-04514-8 978-0-262-36624-3.

\bibitem[Narayanan and Kapoor(2023{\natexlab{a}})]{narayanan2023}
Arvind Narayanan and Sayash Kapoor.
\newblock {GPT}-4 and professional benchmarks: the wrong answer to the wrong question, March 2023{\natexlab{a}}.
\newblock URL \url{https://www.aisnakeoil.com/p/gpt-4-and-professional-benchmarks}.

\bibitem[Narayanan and Kapoor(2023{\natexlab{b}})]{narayanan2023a}
Arvind Narayanan and Sayash Kapoor.
\newblock Evaluating {{LLMs}} is a minefield, 2023{\natexlab{b}}.
\newblock URL \url{https://www.cs.princeton.edu/~arvindn/talks/evaluating_llms_minefield/}.

\bibitem[Nasr et~al.(2023{\natexlab{a}})Nasr, Carlini, Hayase, Jagielski, Cooper, Ippolito, Choquette-Choo, Wallace, and Lee]{nasr2023a}
Milad Nasr, Nicholas Carlini, Jonathan Hayase, Matthew Jagielski, A.~Feder Cooper, Daphne Ippolito, Christopher~A. Choquette-Choo, Eric Wallace, and Katherine Lee.
\newblock Extracting {Training} {Data} from {ChatGPT}, November 2023{\natexlab{a}}.
\newblock URL \url{https://not-just-memorization.github.io/extracting-training-data-from-chatgpt.html?ref=404media.co}.

\bibitem[Nasr et~al.(2023{\natexlab{b}})Nasr, Carlini, Hayase, Jagielski, Cooper, Ippolito, Choquette-Choo, Wallace, Tramèr, and Lee]{nasr2023}
Milad Nasr, Nicholas Carlini, Jonathan Hayase, Matthew Jagielski, A.~Feder Cooper, Daphne Ippolito, Christopher~A. Choquette-Choo, Eric Wallace, Florian Tramèr, and Katherine Lee.
\newblock Scalable {Extraction} of {Training} {Data} from ({Production}) {Language} {Models}, November 2023{\natexlab{b}}.
\newblock URL \url{http://arxiv.org/abs/2311.17035}.

\bibitem[Noroozian(2020)]{noroozian_evaluating_2020}
Arman Noroozian.
\newblock \emph{Evaluating {Hosting} {Provider} {Security} {Through} {Abuse} {Data} and the {Creation} of {Metrics}}.
\newblock Dissertation ({TU} {Delft}), 2020.
\newblock ISBN: 9789065624451.

\bibitem[Oakden-Rayner et~al.(2019)Oakden-Rayner, Dunnmon, Carneiro, and Ré]{oakden-rayner2019}
Luke Oakden-Rayner, Jared Dunnmon, Gustavo Carneiro, and Christopher Ré.
\newblock Hidden {Stratification} {Causes} {Clinically} {Meaningful} {Failures} in {Machine} {Learning} for {Medical} {Imaging}, November 2019.
\newblock URL \url{http://arxiv.org/abs/1909.12475}.

\bibitem[Ojewale et~al.(2024)Ojewale, Steed, Vecchione, Birhane, and Raji]{ojewale2024}
Victor Ojewale, Ryan Steed, Briana Vecchione, Abeba Birhane, and Inioluwa~Deborah Raji.
\newblock Towards {AI} {Accountability} {Infrastructure}: {Gaps} and {Opportunities} in {AI} {Audit} {Tooling}, March 2024.
\newblock URL \url{http://arxiv.org/abs/2402.17861}.

\bibitem[Orr and Crawford(2024{\natexlab{a}})]{orr2024}
Will Orr and Kate Crawford.
\newblock The social construction of datasets: {On} the practices, processes, and challenges of dataset creation for machine learning.
\newblock \emph{New Media \& Society}, 26\penalty0 (9):\penalty0 4955--4972, 2024{\natexlab{a}}.

\bibitem[Orr and Crawford(2024{\natexlab{b}})]{orr2024b}
Will Orr and Kate Crawford.
\newblock Building {Better} {Datasets}: {Seven} {Recommendations} for {Responsible} {Design} from {Dataset} {Creators}, August 2024{\natexlab{b}}.
\newblock URL \url{http://arxiv.org/abs/2409.00252}.

\bibitem[Orr and Kang(2024)]{orr2024a}
Will Orr and Edward~B. Kang.
\newblock {AI} as a {Sport}: {On} the {Competitive} {Epistemologies} of {Benchmarking}.
\newblock In \emph{The 2024 {ACM} {Conference} on {Fairness}, {Accountability}, and {Transparency}}, pages 1875--1884, Rio de Janeiro Brazil, June 2024. ACM.
\newblock ISBN 9798400704505.
\newblock \doi{10.1145/3630106.3659012}.
\newblock URL \url{https://dl.acm.org/doi/10.1145/3630106.3659012}.

\bibitem[Ott et~al.(2022)Ott, Barbosa-Silva, Blagec, Brauner, and Samwald]{ott2022}
Simon Ott, Adriano Barbosa-Silva, Kathrin Blagec, Jan Brauner, and Matthias Samwald.
\newblock Mapping global dynamics of benchmark creation and saturation in artificial intelligence.
\newblock \emph{Nature Communications}, 13\penalty0 (1):\penalty0 6793, November 2022.
\newblock ISSN 2041-1723.
\newblock \doi{10.1038/s41467-022-34591-0}.
\newblock URL \url{https://www.nature.com/articles/s41467-022-34591-0}.

\bibitem[{Oxford English Dictionary}(2017)]{oed}
{Oxford English Dictionary}.
\newblock Benchmark. meaning and use, 2017.
\newblock URL \url{https://www.oed.com/dictionary/benchmark_n?tab=meaning_and_use&tl=true}.

\bibitem[Pacchiardi et~al.(2024)Pacchiardi, Tesic, Cheke, and Hernández-Orallo]{pacchiardi2024}
Lorenzo Pacchiardi, Marko Tesic, Lucy~G. Cheke, and José Hernández-Orallo.
\newblock Leaving the barn door open for {Clever} {Hans}: {Simple} features predict {LLM} benchmark answers, October 2024.
\newblock URL \url{http://arxiv.org/abs/2410.11672}.

\bibitem[Park and Jeoung(2022)]{park2022}
Jaihyun Park and Sullam Jeoung.
\newblock Raison d’être of the benchmark dataset: {A} {Survey} of {Current} {Practices} of {Benchmark} {Dataset} {Sharing} {Platforms}.
\newblock In \emph{Proceedings of {NLP} {Power}! {The} {First} {Workshop} on {Efficient} {Benchmarking} in {NLP}}, pages 1--10, Dublin, Ireland, 2022. Association for Computational Linguistics.
\newblock \doi{10.18653/v1/2022.nlppower-1.1}.
\newblock URL \url{https://aclanthology.org/2022.nlppower-1.1}.

\bibitem[Paullada et~al.(2021)Paullada, Raji, Bender, Denton, and Hanna]{paullada2021}
Amandalynne Paullada, Inioluwa~Deborah Raji, Emily~M. Bender, Emily Denton, and Alex Hanna.
\newblock Data and its (dis)contents: {A} survey of dataset development and use in machine learning research.
\newblock \emph{Patterns}, 2\penalty0 (11):\penalty0 100336, November 2021.
\newblock ISSN 26663899.
\newblock \doi{10.1016/j.patter.2021.100336}.
\newblock URL \url{http://arxiv.org/abs/2012.05345}.

\bibitem[Pfister and Jud(2025)]{pfister2025}
Rolf Pfister and Hansueli Jud.
\newblock Understanding and {Benchmarking} {Artificial} {Intelligence}: {OpenAI}'s o3 {Is} {Not} {AGI}, January 2025.
\newblock URL \url{http://arxiv.org/abs/2501.07458}.

\bibitem[Pinch(1993)]{pinch1993}
Trevor Pinch.
\newblock "{Testing} - {One}, {Two}, {Three} ... {Testing}!": {Toward} a {Sociology} of {Testing}.
\newblock \emph{Science, Technology, \& Human Values}, 18\penalty0 (1):\penalty0 25--41, January 1993.
\newblock ISSN 0162-2439, 1552-8251.
\newblock \doi{10.1177/016224399301800103}.
\newblock URL \url{https://journals.sagepub.com/doi/10.1177/016224399301800103}.

\bibitem[Poelman and Lhoneux(2024)]{poelman_roles_2024}
Wessel Poelman and Miryam~de Lhoneux.
\newblock The {Roles} of {English} in {Evaluating} {Multilingual} {Language} {Models}, December 2024.
\newblock URL \url{http://arxiv.org/abs/2412.08392}.

\bibitem[Qi et~al.(2023)Qi, Zeng, Xie, Chen, Jia, Mittal, and Henderson]{qi2023}
Xiangyu Qi, Yi~Zeng, Tinghao Xie, Pin-Yu Chen, Ruoxi Jia, Prateek Mittal, and Peter Henderson.
\newblock Fine-tuning {Aligned} {Language} {Models} {Compromises} {Safety}, {Even} {When} {Users} {Do} {Not} {Intend} {To}!, October 2023.
\newblock URL \url{http://arxiv.org/abs/2310.03693}.

\bibitem[Raji et~al.(2021)Raji, Denton, Bender, Hanna, and Paullada]{raji2021}
Inioluwa~Deborah Raji, Emily Denton, Emily~M. Bender, Alex Hanna, and Amandalynne Paullada.
\newblock {AI} and the everything in the whole wide world benchmark.
\newblock In \emph{Thirty-fifth Conference on Neural Information Processing Systems Datasets and Benchmarks Track (Round 2)}, 2021.
\newblock URL \url{https://openreview.net/forum?id=j6NxpQbREA1}.

\bibitem[Rauh et~al.(2024)Rauh, Marchal, Manzini, Hendricks, Comanescu, Akbulut, Stepleton, Mateos-Garcia, Bergman, Kay, Griffin, Bariach, Gabriel, Rieser, Isaac, and Weidinger]{rauh2024}
Maribeth Rauh, Nahema Marchal, Arianna Manzini, Lisa~Anne Hendricks, Ramona Comanescu, Canfer Akbulut, Tom Stepleton, Juan Mateos-Garcia, Stevie Bergman, Jackie Kay, Conor Griffin, Ben Bariach, Iason Gabriel, Verena Rieser, William Isaac, and Laura Weidinger.
\newblock Gaps in the {Safety} {Evaluation} of {Generative} {AI}.
\newblock \emph{Proceedings of the AAAI/ACM Conference on AI, Ethics, and Society}, 7:\penalty0 1200--1217, October 2024.
\newblock ISSN 3065-8365.
\newblock \doi{10.1609/aies.v7i1.31717}.
\newblock URL \url{https://ojs.aaai.org/index.php/AIES/article/view/31717}.

\bibitem[Ren et~al.(2024)Ren, Basart, Khoja, Gatti, Phan, Yin, Mazeika, Pan, Mukobi, Kim, Fitz, and Hendrycks]{ren2024}
Richard Ren, Steven Basart, Adam Khoja, Alice Gatti, Long Phan, Xuwang Yin, Mantas Mazeika, Alexander Pan, Gabriel Mukobi, Ryan~Hwang Kim, Stephen Fitz, and Dan Hendrycks.
\newblock Safetywashing: Do {AI} safety benchmarks actually measure safety progress?
\newblock In \emph{The Thirty-eight Conference on Neural Information Processing Systems Datasets and Benchmarks Track}, 2024.
\newblock URL \url{https://openreview.net/forum?id=YagfTP3RK6}.

\bibitem[Reuel et~al.(2024)Reuel, Hardy, Smith, Lamparth, Hardy, and Kochenderfer]{reuel_betterbench_2024}
Anka Reuel, Amelia Hardy, Chandler Smith, Max Lamparth, Malcolm Hardy, and Mykel Kochenderfer.
\newblock Betterbench: Assessing {AI} benchmarks, uncovering issues, and establishing best practices.
\newblock In \emph{The Thirty-eight Conference on Neural Information Processing Systems Datasets and Benchmarks Track}, 2024.
\newblock URL \url{https://openreview.net/forum?id=hcOq2buakM}.

\bibitem[Roberts et~al.(2023)Roberts, Thakur, Herlihy, White, and Dooley]{roberts2023}
Manley Roberts, Himanshu Thakur, Christine Herlihy, Colin White, and Samuel Dooley.
\newblock Data {Contamination} {Through} the {Lens} of {Time}, October 2023.
\newblock URL \url{http://arxiv.org/abs/2310.10628}.

\bibitem[Rodriguez et~al.(2021)Rodriguez, Barrow, Hoyle, Lalor, Jia, and Boyd-Graber]{rodriguez2021}
Pedro Rodriguez, Joe Barrow, Alexander~Miserlis Hoyle, John~P. Lalor, Robin Jia, and Jordan Boyd-Graber.
\newblock Evaluation {Examples} are not {Equally} {Informative}: {How} should that change {NLP} {Leaderboards}?
\newblock In \emph{Proceedings of the 59th {Annual} {Meeting} of the {Association} for {Computational} {Linguistics} and the 11th {International} {Joint} {Conference} on {Natural} {Language} {Processing} ({Volume} 1: {Long} {Papers})}, pages 4486--4503, Online, 2021. Association for Computational Linguistics.
\newblock \doi{10.18653/v1/2021.acl-long.346}.
\newblock URL \url{https://aclanthology.org/2021.acl-long.346}.

\bibitem[Roose(2024)]{roose2024}
Kevin Roose.
\newblock A.i. has a measurement problem.
\newblock Report, New York Times, 2024.
\newblock URL \url{https://www.nytimes.com/2024/04/15/technology/ai-models-measurement.html}.

\bibitem[Röttger et~al.(2024)Röttger, Pernisi, Vidgen, and Hovy]{rottger2024}
Paul Röttger, Fabio Pernisi, Bertie Vidgen, and Dirk Hovy.
\newblock {SafetyPrompts}: a {Systematic} {Review} of {Open} {Datasets} for {Evaluating} and {Improving} {Large} {Language} {Model} {Safety}, April 2024.
\newblock URL \url{http://arxiv.org/abs/2404.05399}.

\bibitem[Sambasivan et~al.(2021)Sambasivan, Kapania, Highfill, Akrong, Paritosh, and Aroyo]{sambasivan2021}
Nithya Sambasivan, Shivani Kapania, Hannah Highfill, Diana Akrong, Praveen Paritosh, and Lora~M Aroyo.
\newblock “{Everyone} wants to do the model work, not the data work”: {Data} {Cascades} in {High}-{Stakes} {AI}.
\newblock In \emph{Proceedings of the 2021 {CHI} {Conference} on {Human} {Factors} in {Computing} {Systems}}, pages 1--15, Yokohama Japan, May 2021. ACM.
\newblock ISBN 978-1-4503-8096-6.
\newblock \doi{10.1145/3411764.3445518}.
\newblock URL \url{https://dl.acm.org/doi/10.1145/3411764.3445518}.

\bibitem[Scheuerman et~al.(2021)Scheuerman, Hanna, and Denton]{scheuerman2021}
Morgan~Klaus Scheuerman, Alex Hanna, and Emily Denton.
\newblock Do {Datasets} {Have} {Politics}? {Disciplinary} {Values} in {Computer} {Vision} {Dataset} {Development}.
\newblock \emph{Proceedings of the ACM on Human-Computer Interaction}, 5\penalty0 (CSCW2):\penalty0 1--37, October 2021.
\newblock ISSN 2573-0142.
\newblock \doi{10.1145/3476058}.
\newblock URL \url{https://dl.acm.org/doi/10.1145/3476058}.

\bibitem[Schlangen(2020)]{schlangen_targeting_2020}
David Schlangen.
\newblock Targeting the {Benchmark}: {On} {Methodology} in {Current} {Natural} {Language} {Processing} {Research}, July 2020.
\newblock URL \url{http://arxiv.org/abs/2007.04792}.

\bibitem[Sculley et~al.(2018)Sculley, Snoek, Rahimi, and Wiltschko]{sculley2018}
D~Sculley, Jasper Snoek, Ali Rahimi, and Alex Wiltschko.
\newblock Winner's {Curse}? {On} {Pace}, {Progress}, and {Empirical} {Rigor}.
\newblock Vancouver, BC, Canada, 2018.
\newblock URL \url{https://openreview.net/pdf?id=rJWF0Fywf}.

\bibitem[Selbst et~al.(2019)Selbst, Boyd, Friedler, Venkatasubramanian, and Vertesi]{selbst2019}
Andrew~D. Selbst, Danah Boyd, Sorelle~A. Friedler, Suresh Venkatasubramanian, and Janet Vertesi.
\newblock Fairness and {Abstraction} in {Sociotechnical} {Systems}.
\newblock In \emph{Proceedings of the {Conference} on {Fairness}, {Accountability}, and {Transparency}}, pages 59--68, Atlanta GA USA, January 2019. ACM.
\newblock ISBN 978-1-4503-6125-5.
\newblock \doi{10.1145/3287560.3287598}.
\newblock URL \url{https://dl.acm.org/doi/10.1145/3287560.3287598}.

\bibitem[Sen et~al.(2015)Sen, Giesel, Gold, Hillmann, Lesicko, Naden, Russell, Wang, and Hecht]{sen2015}
Shilad Sen, Margaret~E. Giesel, Rebecca Gold, Benjamin Hillmann, Matt Lesicko, Samuel Naden, Jesse Russell, Zixiao~(Ken) Wang, and Brent Hecht.
\newblock Turkers, {Scholars}, "{Arafat}" and "{Peace}": {Cultural} {Communities} and {Algorithmic} {Gold} {Standards}.
\newblock In \emph{Proceedings of the 18th {ACM} {Conference} on {Computer} {Supported} {Cooperative} {Work} \& {Social} {Computing}}, pages 826--838, Vancouver BC Canada, February 2015. ACM.
\newblock ISBN 978-1-4503-2922-4.
\newblock \doi{10.1145/2675133.2675285}.
\newblock URL \url{https://dl.acm.org/doi/10.1145/2675133.2675285}.

\bibitem[Simson et~al.(2024)Simson, Fabris, and Kern]{simson2024}
Jan Simson, Alessandro Fabris, and Christoph Kern.
\newblock Lazy {Data} {Practices} {Harm} {Fairness} {Research}.
\newblock In \emph{The 2024 {ACM} {Conference} on {Fairness}, {Accountability}, and {Transparency}}, pages 642--659, Rio de Janeiro Brazil, June 2024. ACM.
\newblock ISBN 9798400704505.
\newblock \doi{10.1145/3630106.3658931}.
\newblock URL \url{https://dl.acm.org/doi/10.1145/3630106.3658931}.

\bibitem[Smith et~al.(2022)Smith, Amershi, Barocas, Wallach, and Vaughan]{smith2022}
Jessie~J. Smith, Saleema Amershi, Solon Barocas, Hanna Wallach, and Jennifer~Wortman Vaughan.
\newblock {REAL} {ML}: {Recognizing}, {Exploring}, and {Articulating} {Limitations} of {Machine} {Learning} {Research}.
\newblock In \emph{2022 {ACM} {Conference} on {Fairness}, {Accountability}, and {Transparency}}, pages 587--597, June 2022.
\newblock \doi{10.1145/3531146.3533122}.
\newblock URL \url{http://arxiv.org/abs/2205.08363}.

\bibitem[Srivastava et~al.(2023)Srivastava, Rastogi, Rao, Shoeb, Abid, Fisch, Brown, Santoro, Gupta, Garriga-Alonso, Kluska, Lewkowycz, Agarwal, Power, Ray, Warstadt, et~al.]{srivastava2023}
Aarohi Srivastava, Abhinav Rastogi, Abhishek Rao, Abu Awal~Md Shoeb, Abubakar Abid, Adam Fisch, Adam~R. Brown, Adam Santoro, Aditya Gupta, Adrià Garriga-Alonso, Agnieszka Kluska, Aitor Lewkowycz, Akshat Agarwal, Alethea Power, Alex Ray, Alex Warstadt, et~al.
\newblock Beyond the {Imitation} {Game}: {Quantifying} and extrapolating the capabilities of language models, June 2023.
\newblock URL \url{http://arxiv.org/abs/2206.04615}.

\bibitem[Stengers(2018)]{stengers2018}
Isabelle Stengers.
\newblock \emph{Another science is possible: a manifesto for slow science}.
\newblock Polity press, Cambridge, 2018.
\newblock ISBN 978-1-5095-2180-7.

\bibitem[Stewart(2023)]{Stewart2023}
Matthew Stewart.
\newblock The olympics of ai: Benchmarking machine learning systems, 2023.
\newblock URL \url{https://towardsdatascience.com/the-olympics-of-ai-benchmarking-machine-learning-systems-c4b2051fbd2b}.

\bibitem[Strathern(1997)]{strathern1997}
Marilyn Strathern.
\newblock ‘{Improving} ratings’: audit in the {British} {University} system.
\newblock \emph{European Review}, 5\penalty0 (3):\penalty0 305--321, July 1997.
\newblock ISSN 10627987, 1234981X.
\newblock \doi{10.1002/(SICI)1234-981X(199707)5:3<305::AID-EURO184>3.0.CO;2-4}.
\newblock URL \url{https://www.cambridge.org/core/product/identifier/S1062798700002660/type/journal_article}.

\bibitem[Subramonian et~al.(2023)Subramonian, Yuan, III, and Blodgett]{subramonian2023}
Arjun Subramonian, Xingdi Yuan, Hal~Daumé III, and Su~Lin Blodgett.
\newblock It {Takes} {Two} to {Tango}: {Navigating} {Conceptualizations} of {NLP} {Tasks} and {Measurements} of {Performance}, May 2023.
\newblock URL \url{http://arxiv.org/abs/2305.09022}.

\bibitem[Thakur et~al.(2021)Thakur, Reimers, R{\"u}ckl{\'e}, Srivastava, and Gurevych]{thakur2021beir}
Nandan Thakur, Nils Reimers, Andreas R{\"u}ckl{\'e}, Abhishek Srivastava, and Iryna Gurevych.
\newblock {BEIR}: A heterogeneous benchmark for zero-shot evaluation of information retrieval models.
\newblock In \emph{Thirty-fifth Conference on Neural Information Processing Systems Datasets and Benchmarks Track (Round 2)}, 2021.
\newblock URL \url{https://openreview.net/forum?id=wCu6T5xFjeJ}.

\bibitem[{The White House}(2023)]{USAIEO2023}
{The White House}.
\newblock {Executive Order on the Safe, Secure, and Trustworthy Development and Use of Artificial Intelligence}, 2023.

\bibitem[Thylstrup et~al.(2022)Thylstrup, Hansen, Flyverbom, and Amoore]{thylstrup2022}
Nanna~Bonde Thylstrup, Kristian~Bondo Hansen, Mikkel Flyverbom, and Louise Amoore.
\newblock Politics of data reuse in machine learning systems: {Theorizing} reuse entanglements.
\newblock \emph{Big Data \& Society}, 9\penalty0 (2):\penalty0 20539517221139785, July 2022.
\newblock ISSN 2053-9517, 2053-9517.
\newblock \doi{10.1177/20539517221139785}.
\newblock URL \url{https://journals.sagepub.com/doi/10.1177/20539517221139785}.

\bibitem[Tirumala et~al.(2022)Tirumala, Markosyan, Zettlemoyer, and Aghajanyan]{tirumala2022}
Kushal Tirumala, Aram Markosyan, Luke Zettlemoyer, and Armen Aghajanyan.
\newblock Memorization {Without} {Overfitting}: {Analyzing} the {Training} {Dynamics} of {Large} {Language} {Models}.
\newblock In S.~Koyejo, S.~Mohamed, A.~Agarwal, D.~Belgrave, K.~Cho, and A.~Oh, editors, \emph{Advances in {Neural} {Information} {Processing} {Systems}}, volume~35, pages 38274--38290. Curran Associates, Inc., 2022.
\newblock URL \url{https://proceedings.neurips.cc/paper_files/paper/2022/file/fa0509f4dab6807e2cb465715bf2d249-Paper-Conference.pdf}.

\bibitem[Tsipras et~al.(2020)Tsipras, Santurkar, Engstrom, Ilyas, and Madry]{tsipras2020}
Dimitris Tsipras, Shibani Santurkar, Logan Engstrom, Andrew Ilyas, and Aleksander Madry.
\newblock From {ImageNet} to {Image} {Classification}: {Contextualizing} {Progress} on {Benchmarks}, May 2020.
\newblock URL \url{http://arxiv.org/abs/2005.11295}.

\bibitem[{UK Parliament}(2023)]{UKSafetyAct2023}
{UK Parliament}.
\newblock {Online Safety Act 2023}, 2023.

\bibitem[{US Department of Commerce}(2025)]{USAIDiff2025}
{US Department of Commerce}.
\newblock {Framework for Artificial Intelligence Diffusion}, 2025.

\bibitem[Vafa et~al.(2024)Vafa, Chen, Rambachan, Kleinberg, and Mullainathan]{vafa2024}
Keyon Vafa, Justin~Y. Chen, Ashesh Rambachan, Jon Kleinberg, and Sendhil Mullainathan.
\newblock Evaluating the {World} {Model} {Implicit} in a {Generative} {Model}, November 2024.
\newblock URL \url{http://arxiv.org/abs/2406.03689}.

\bibitem[Weidinger et~al.(2023)Weidinger, Rauh, Marchal, Manzini, Hendricks, Mateos-Garcia, Bergman, Kay, Griffin, Bariach, Gabriel, Rieser, and Isaac]{weidinger_sociotechnical_2023}
Laura Weidinger, Maribeth Rauh, Nahema Marchal, Arianna Manzini, Lisa~Anne Hendricks, Juan Mateos-Garcia, Stevie Bergman, Jackie Kay, Conor Griffin, Ben Bariach, Iason Gabriel, Verena Rieser, and William Isaac.
\newblock Sociotechnical {Safety} {Evaluation} of {Generative} {AI} {Systems}, October 2023.
\newblock URL \url{http://arxiv.org/abs/2310.11986}.

\bibitem[Weij et~al.(2024)Weij, Hofstätter, Jaffe, Brown, and Ward]{weij2024}
Teun van~der Weij, Felix Hofstätter, Ollie Jaffe, Samuel~F. Brown, and Francis~Rhys Ward.
\newblock {AI} {Sandbagging}: {Language} {Models} can {Strategically} {Underperform} on {Evaluations}, June 2024.
\newblock URL \url{http://arxiv.org/abs/2406.07358}.

\bibitem[Xu et~al.(2024{\natexlab{a}})Xu, Guan, Greene, and Kechadi]{xu2024}
Cheng Xu, Shuhao Guan, Derek Greene, and M.-Tahar Kechadi.
\newblock Benchmark {Data} {Contamination} of {Large} {Language} {Models}: {A} {Survey}, June 2024{\natexlab{a}}.
\newblock URL \url{http://arxiv.org/abs/2406.04244}.

\bibitem[Xu et~al.(2024{\natexlab{b}})Xu, Wang, Fan, and Liu]{xu_benchmarking_2024}
Ruijie Xu, Zengzhi Wang, Run-Ze Fan, and Pengfei Liu.
\newblock Benchmarking {Benchmark} {Leakage} in {Large} {Language} {Models}, April 2024{\natexlab{b}}.
\newblock URL \url{http://arxiv.org/abs/2404.18824}.

\bibitem[Yang et~al.(2023)Yang, Chiang, Zheng, Gonzalez, and Stoica]{yang_rethinking_2023}
Shuo Yang, Wei-Lin Chiang, Lianmin Zheng, Joseph~E. Gonzalez, and Ion Stoica.
\newblock Rethinking {Benchmark} and {Contamination} for {Language} {Models} with {Rephrased} {Samples}, November 2023.
\newblock URL \url{http://arxiv.org/abs/2311.04850}.

\bibitem[Yuan et~al.(2023)Yuan, Chen, Cui, Gao, Zou, Cheng, Ji, Liu, and Sun]{yuan2023}
Lifan Yuan, Yangyi Chen, Ganqu Cui, Hongcheng Gao, Fangyuan Zou, Xingyi Cheng, Heng Ji, Zhiyuan Liu, and Maosong Sun.
\newblock Revisiting {Out}-of-distribution {Robustness} in {NLP}: {Benchmark}, {Analysis}, and {LLMs} {Evaluations}.
\newblock \emph{37th Conference on Neural Information Processing Systems (NeurIPS 2023) Track on Datasets and Benchmarks}, 2023.

\bibitem[Zhang et~al.(2024)Zhang, Klyman, Mai, Levine, Zhang, Bommasani, and Liang]{zhang2024}
Andy~K. Zhang, Kevin Klyman, Yifan Mai, Yoav Levine, Yian Zhang, Rishi Bommasani, and Percy Liang.
\newblock Language model developers should report train-test overlap, October 2024.
\newblock URL \url{http://arxiv.org/abs/2410.08385}.

\bibitem[Zhijia(2024)]{Zhijia2024}
Lin Zhijia.
\newblock Top llms in china and the u.s. only 5 months apart: Kai-fu lee, 2024.
\newblock URL \url{https://en.tmtpost.com/post/7289212}.

\end{thebibliography}

\end{document}